\documentclass[acmsmall]{acmart}
\makeatletter
\def\runningfoot{\def\@runningfoot{}}
\def\firstfoot{\def\@firstfoot{}}
\makeatother
\renewcommand\footnotetextcopyrightpermission[1]{} % removes footnote with conference information in first column
\usepackage{fancyhdr}
\usepackage{xspace}

\usepackage{bbm}

\makeatletter
\DeclareRobustCommand\onedot{\futurelet\@let@token\@onedot}
\def\@onedot{\ifx\@let@token.\else.\null\fi\xspace}

\def\eg{\textit{e.g}\onedot} 
\def\ie{\textit{i.e}\onedot} 
\def\cf{\textit{c.f}\onedot}

\def\etal{\textit{et al}\onedot}
\makeatother
\usepackage{booktabs}
\usepackage{multirow}
\usepackage{xcolor}
\usepackage{adjustbox}
\usepackage{threeparttable}
\usepackage{bm}

\usepackage{amsfonts,amssymb}
\usepackage{ulem}
\usepackage{epsfig}
\usepackage{subfig}
\usepackage{graphicx}
\usepackage{amsmath}
\usepackage{enumitem} % for indent in itemize

\AtBeginDocument{%
  \providecommand\BibTeX{{%
    \normalfont B\kern-0.5em{\scshape i\kern-0.25em b}\kern-0.8em\TeX}}}

\usepackage{numprint}
\usepackage{wrapfig}
\usepackage{ulem}

\begin{document}

\title{A Closer Look at Debiased Temporal Sentence Grounding in Videos: Dataset, Metric, and Approach}

\author{Xiaohan Lan}
% \authornote{Partial work done while Xiaohan Lan was a research intern at Meituan.}
\email{lanxh20@mails.tsinghua.edu.cn}
\affiliation{
  \institution{Tsinghua Shenzhen International Graduate School}
  \streetaddress{University Town of Shenzhen}
  \city{Shenzhen}
  \country{China}}

\author{Yitian Yuan}
\affiliation{
  \institution{Meituan}
  \city{Beijing}
  \country{China}}
\email{yuanyitian@foxmail.com}

\author{Xin Wang}
\affiliation{
  \institution{Tsinghua University}
  \streetaddress{30 Shuangqing Rd}
  \city{Beijing}
  \country{China}}
 \email{xin_wang@tsinghua.edu.cn}
 
\author{Long Chen}
\authornote{Corresponding authors.}
\email{zjuchenlong@gmail.com}
\affiliation{
  \institution{Columbia University}
  \city{New York}
  \country{USA}
}

\author{Zhi Wang}
\affiliation{
  \institution{Tsinghua Shenzhen International Graduate School}
  \streetaddress{University Town of Shenzhen}
  \city{Shenzhen}
  \country{China}}
 \email{wangzhi@sz.tsinghua.edu.cn}

\author{Lin Ma}
\affiliation{
  \institution{Meituan}
  \city{Beijing}
  \country{China}}
\email{forest.linma@gmail.com}

\author{Wenwu Zhu}
\authornotemark[1]
\affiliation{
  \institution{Tsinghua University}
  \streetaddress{30 Shuangqing Rd}
  \city{Beijing}
  \country{China}}
 \email{wwzhu@tsinghua.edu.cn}
 
\renewcommand{\shortauthors}{Lan, et al.}

\begin{abstract}
Temporal Sentence Grounding in Videos (TSGV), which aims to ground a natural language sentence that indicates complex human activities in an untrimmed video, has drawn widespread attention over the past few years. However, recent studies have found that current benchmark datasets may have obvious moment annotation biases, enabling several simple baselines even without training to achieve state-of-the-art (SOTA) performance. In this paper, we take a closer look at existing evaluation protocols for TSGV, and find that both the prevailing dataset splits and evaluation metrics are the devils that lead to untrustworthy benchmarking. Therefore, we propose to re-organize the two widely-used datasets, making the ground-truth moment distributions different in the training and test splits, \ie, out-of-distribution (OOD) test. Meanwhile, we introduce a new evaluation metric ``dR@$n$,IoU@$m$'' that discounts the basic recall scores especially with small IoU thresholds, so as to alleviate the inflating evaluation caused by biased datasets with a large proportion of long ground-truth moments. New benchmarking results indicate that our proposed evaluation protocols can better monitor the research progress in TSGV. Furthermore, we propose a novel causality-based Multi-branch Deconfounding Debiasing (MDD) framework for unbiased moment prediction. Specifically, we design a multi-branch deconfounder to eliminate the effects caused by multiple confounders with causal intervention. In order to help the model better align the semantics between sentence queries and video moments, we enhance the representations during feature encoding. Specifically, for textual information, the query is parsed into several verb-centered phrases to obtain a more fine-grained textual feature. For visual information, the positional information has been decomposed from the moment features to enhance the representations of moments with diverse locations. Extensive experiments demonstrate that our proposed approach can achieve competitive results among existing SOTA approaches and outperform the base model with great gains.
\end{abstract}



\maketitle

\section{Introduction}

\begin{figure}[!t]
	\centering
	\setlength{\belowcaptionskip}{-0.2cm}
	\includegraphics[width=0.9\columnwidth]{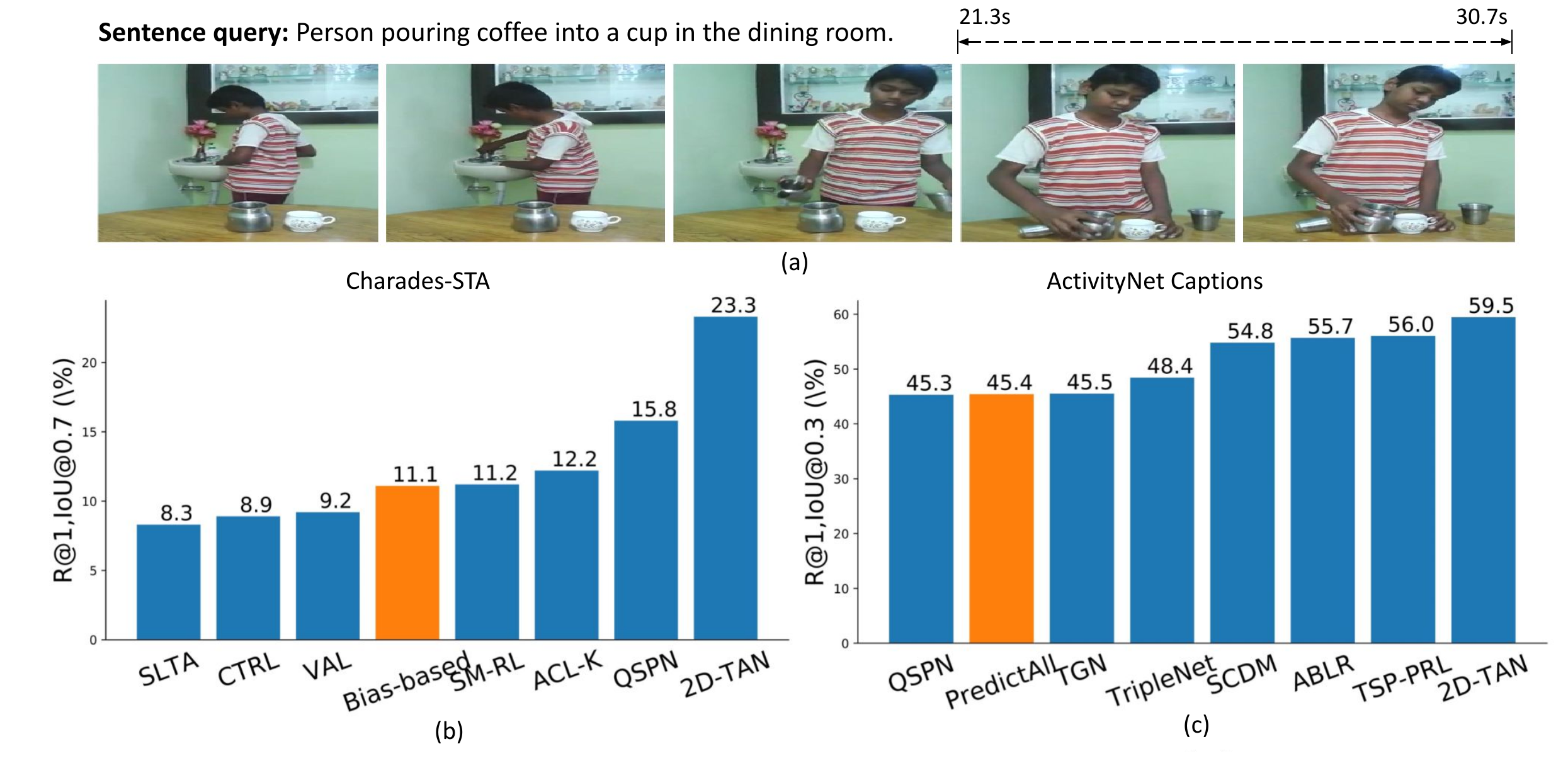}
	\caption{(a): Given an untrimmed video and a sentence query, TSGV aims to localize the semantic-related moment with the start timestamp (21.3s) and end timestamp (30.7s). (b): The performance comparisons of some SOTA TSGV models with the \texttt{Bias-based} baseline (orange bar) on Charades-STA with evaluation metric ``R@$1$,IoU@$0.7$''. (c): The performance comparisons of some SOTA TSGV models with \texttt{PredictAll} baseline (orange bar) on ActivityNet Captions with evaluation metric ``R@$1$,IoU@$0.3$''.
	}
	\label{fig:intro}
\end{figure}

Temporal Sentence Grounding in Videos (TSGV) has received increased attention in recent years. Specifically, given one descriptive sentence, the TSGV task aims to retrieve a video segment (\ie, moment) from an untrimmed video corresponding to the sentence query. For example, as shown in Fig.~\ref{fig:intro} (a), when the sentence describes a person pouring coffee into a cup in the dining room, the corresponding video segment (21.3s-30.7s) should be located. It can be observed that TSGV needs to understand both visual information in videos and textual information in sentences, which is an extremely challenging task in the multimedia community~\cite{zhu2015multimedia,peng2017cross}.

In recent years, a number of approaches~\cite{gao2017tall,anne2017localizing,yuan2019find,chen2018temporally,zhang2020learning,mithun2019weakly,wu2020tree} have emerged to solve the TSGV problem. Although each newly proposed method can plausibly achieve better performance than the previous one, a recent study~\cite{otani2020uncovering} reveals that current state-of-the-art (SOTA) methods may take shortcuts by fitting the ground-truth moment annotation distribution biases, without truly understanding the multimodal inputs. As shown in the Fig.~\ref{fig:intro} (b), the \texttt{Bias-based} approach, which samples a moment from the frequency statistics of the ground-truth moment annotations in the training set as prediction, can unexpectedly outperform several SOTA deep models on Charades-STA~\cite{gao2017tall} dataset. This observation indicates that current benchmark datasets may have obvious biases in terms of moment location distribution, and it is hard to judge whether existing methods are merely fitting the biases or truly learning the semantic alignment relationship between the two modalities. Another characteristic of biased datasets is that they have a large proportion of long samples, \eg, 40\% queries in the ActivityNet Captions dataset~\cite{krishna2017dense} refer to a moment occupying over 30\% temporal ranges of the whole input video. Since prevailing metric for TSGV task is ``R@$n$,IoU@$m$'', \ie, the percentage of testing samples which have at least one of the top-$n$ results with IoU larger than $m$, these overlong ground-truth moments can be hit easily especially with small IoU threshold $m$, resulting in untrustworthy evaluation results. As an extreme case, a simple baseline which directly returns the whole video as the prediction (\cf, the \texttt{PredictAll} baseline in Fig.\ref{fig:intro} (c)) can still achieve a SOTA performance with the metric of ``R@$1$,IoU@$0.3$''.

Therefore, to disentangle the effect caused by the biases and alleviate the inflating evaluation, we propose to re-split the datasets and design a new metric. Specifically, we re-organize two widely-used datasets, \ie, Charades-STA and ActivityNet Captions and name them \textbf{Charades-CD} and \textbf{ActivityNet-CD} (CD means under Changing Distribution). For each dataset, besides the test set with the same distribution as the training set (test-iid), we also construct a test set with a completely different distribution of moment locations from the training set (dubbed as test-ood set), \ie, Out-Of-Distribution (OOD) test. As for metrics, we design a new evaluation metric ``dR@$n$,IoU@$m$'' that takes temporal distances between the predicted moment and ground-truth moment into consideration. The new metric can discount the basic recall scores especially under small IoU thresholds. So our proposed evaluation protocols (\ie, re-organized dataset splits and improved evaluation metric) are able to provide more trustworthy evaluation results for existing methods and figure out whether they just fit the moment annotation biases. Several representative TSGV methods are tested with such new evaluation protocols and one key finding is that the performances of the vast majority of these methods degrade significantly on test-ood set compared to test-iid set, which indicates the fact that current methods heavily rely on the biases for moment prediction. Therefore, how to effectively debias a TSGV model to make it truly focus on the semantic alignment between the two modalities becomes one primary issue to be addressed.

% \textcolor{red}{Add evaluation metrics and make the datasets with name Charades-CD and ActivityNet-CD.}

% \textcolor{red}{Currently, there is a proliferation of debiasing methods based on re-sampling or re-weighting with good results, but they are weak in interpretability. The subsequent emergence of causality-based debiasing methods can both explain the reasons why the above methods can be effective based on the theory of causality and propose a new framework from the perspective of causal intervention or counterfactual reasoning.}
To reduce the effects of moment annotation biases, we further propose a novel causality-based \textbf{Multi-branch Deconfounding Debiasing} (MDD) framework. Specifically, by constructing a causal graph, we find that multiple confounders can lead to spurious correlations between the multimodal inputs (video moments and descriptive sentences) and final predicted matching scores. Thus we present a multi-branch deconfounder to block the effects caused by the confounders with backdoor adjustment. Furthermore, we also enhance the representation capability for two modalities. For textual information, we exploit a semantic role labeling toolkit to parse the sentence into a three-layer semantic role tree, and a more fine-grained sentence feature is obtained by adopting hierarchical attention mechanism on the tree. For visual information, in order to discriminate video moments and distinguish different temporal relationships, a reconstruction loss function is created to enhance the video moment features. Extensive experiments demonstrate that the adopting of our debiasing strategy can significantly improve the grounding accuracy on both the test-iid and test-ood sets of two datasets.

This paper is a substantial extension of our ACM Multimedia HUMA Workshop paper~\cite{yuan2021closer}, which won the best paper award. Compared with the previous version, we make several improvements:
\begin{itemize}[leftmargin=1em]
% \vspace{-1em}
% 	\item We design new evaluation protocols for TSGV, including new splits of two prevailing TSGV datasets and a new metric: dR@$n$,IoU@$m$. The new splits are able to disentangle the effects of annotation biases and the new metric is more reliable than the existing metrics, especially when IoU threshold is small.

	% \item We propose a new metric dRecall@$m$,IoU@$n$, which can reduce the unreliable and inflating evaluation scores caused by temporal duration characteristics of annotated segments, and better evaluate the temporal sentence grounding accuracy.
	
% 	\item \textcolor{red}{(Merge, a model with model name, and some specific model components.)}
    \item we propose a new Multi-branch Deconfounding Debiasing (MDD) framework for unbiased moment prediction. The proposed multi-branch deconfounder can simultaneously remove the spurious correlation between multimodal inputs and predicted scores caused by multiple confounders, avoiding the excessive abuse of dataset biases.
    
    \item We also enhance the visual and textual features for better cross-modal matching. The gated fine-grained feature extractor for queries and position reconstruction module for video moments can capture richer and more discriminative representations of these two modalities.
	
	\item We conduct extensive studies on MDD framework compared with existing SOTA models. The experimental results demonstrate that our approach can achieve competitive results among existing SOTA models and outperform the base model with great gains on both test-iid and test-ood sets.
	
% 	on both test-iid and test-ood sets outperform the baseline model with large gains.
	
% 	Consistent performance gaps between IID and OOD samples have proven that our new evaluation protocols can better monitor the progress in TSGV,
	
	% \item We thoroughly evaluate several temporal sentence grounding methods on both the origin and our created new splits. The results show that most of the existing methods will drop significantly when evaluated on the test-ood set. Our findings demonstrate that it is necessary to rethink the current progress made on the temporal sentence grounding problem, and our proposed splits can provide a good test bed for the debiased temporal grounding.
\end{itemize}

\section{Related Work}

\subsection{Temporal Sentence Grounding in Videos}
Existing TSGV methods can be summarized into four main categories:

\textbf{\textit{Two-Stage Methods.}} Early methods commonly address the TSGV task in a two-stage manner. In particular, they first extract a large number of moment candidates via sliding window sampling strategy, and then either project the query and these candidates into a common space for subsequent cross-modal matching~\cite{anne2017localizing} or fuse the query feature and video moment features to predict the alignment score and refine the moments with position offset regression~\cite{gao2017tall,ge2019mac,jiang2019cross,liu2018attentive,liu2018cross,song2018val,xiao2021boundary,xiao2021natural}. To reduce the number of candidates for accelerating the localization process, Xu~\etal~\cite{xu2019multilevel} proposed QSPN, which filters unlikely video moments by injecting the textual feature into the early process of candidate generation.
% \noindent\textbf{\textit{Two-Stage Methods.}} Early TSGV methods typically solve this problem in a two-stage fashion: They first extract numerous video segment candidates by temporal sliding windows, and then either match the query sentence with these candidates~\cite{anne2017localizing} or fuse query and video segment features to regress the final position, \eg, CTRL~\cite{gao2017tall}, ACL-K~\cite{ge2019mac}, SLTA~\cite{jiang2019cross}, ACRN~\cite{liu2018attentive}, ROLE~\cite{liu2018cross}, VAL~\cite{song2018val} and BPNet~\cite{xiao2021boundary}. To speed up the sliding window processing, Xu~\etal~\cite{xu2019multilevel} proposed QSPN, which injects text features early to generate segment candidates, and helps to eliminate the unlikely segment candidates and increases the grounding accuracy.

\textbf{\textit{End-to-End Methods.}} Other than adopting the two-stage framework that is inefficient due to the redundant computation with pre-segmented overlapping candidate moments, some studies start to address the TSGV task in an end-to-end pipeline~\cite{chen2018temporally,chen2020rethinking,lu2019debug,yuan2019semantic,yuan2019find,zeng2020dense,zhang2019man,zhang2020learning,zhang2019cross,cao2021pursuit}. TGN~\cite{chen2018temporally} adopts LSTM~\cite{hochreiter1997long} to sequentially score a bunch of multi-scale moment candidates ended at each time step in one single pass. Instead of candidate moment scoring, ABLR~\cite{yuan2019find} directly regresses the start and end timestamps of the predicted moments from the attention weights yielded by the multi-turn cross-modal interaction. It is worth noting that both TGN and ABLR use LSTM to process the video stream and some other TSGV frameworks adopt temporal convolutional networks as the solutions. MAN~\cite{zhang2019man} employs a hierarchical convolutional network to encode the whole video stream, where the language features are integrated as its dynamic filters to address semantic misalignment. Yuan~\etal~\cite{yuan2019semantic} presented SCDM, which conducts a query semantics-guided feature normalization process among different temporal convolutional layers. Both MAN and SCDM encode the video sequence with 1D feature map which can naturally indicate the temporal locations and scales of different moments, while 2D-TAN~\cite{zhang2020learning} models the temporal relations between video moments with a 2D temporal map. The 2D temporal map can encode the temporally adjacent relations of diverse moments indicated by their 2D position coordinates. Thus more discriminative moment representations can be learned for cross-modal matching.
% \noindent\textbf{\textit{End-to-End Methods.}} Besides the two-stage framework, some other TSGV works seek to solve the grounding problem in an end-to-end manner~\cite{chen2018temporally,yuan2019semantic,yuan2019find,zeng2020dense,zhang2019man,zhang2020learning,zhang2019cross}. Chen~\etal~\cite{chen2018temporally} proposed TGN, which sequentially scores a set of temporal candidates ended at each frame and generates the final grounding result in one single pass. Similarly, ABLR model also processes video sequences via LSTMs~\cite{yuan2019find}, where the start and end timestamps of the predicted segments are regressed from the attention weights yielded by the multi-pass interaction between videos and queries. There are also some works leveraging temporal convolutional networks to solve the TSGV problem. Zhang~\etal~\cite{zhang2019man} presented MAN, which assigns candidate segment representations aligned with language semantics over different temporal locations and scales in hierarchical temporal convolutional feature maps. Yuan~\etal~\cite{yuan2019semantic} introduced the SCDM, where query semantic is used to control the feature normalization between different temporal convolutional layers, making the query-related video activities tightly compose together. Both MAN and SCDM only consider 1D temporal feature maps, while 2D-TAN~\cite{zhang2020learning} models the temporal relations between video segments by a 2D map. In the 2D map, 2D-TAN encodes the adjacent temporal relation, and learns discriminative features for matching video segments with queries.

\textbf{\textit{RL-based Methods.}} Some recent works employ Reinforcement Learning (RL)-based frameworks, which formulate the TSGV task as a problem of sequential decision making, progressively adjusting the temporal boundaries of predicted moment~\cite{he2019read,wang2019language,hahn2019tripping,wu2020tree}. Specifically, He~\etal~\cite{he2019read} proposed the RL model that iteratively regulates current locations according to the learned policy. The policy network is implemented by a recurrent neural network (RNN) that outputs the probability distribution over its action space. Wang~\etal~\cite{wang2019language} presented a semantic matching RL (SM-RL) model, which is also based on RNN. The SM-RL integrates visual semantic concepts into the video features to bridge the semantic gap between visual and textual information. TripNet~\cite{hahn2019tripping} can efficiently localize the desired moment without watching the entire video, by making the agent learn how to intelligently move the candidate window around the video. Inspired by the coarse-to-fine human decision-making paradigm, Wu~\etal~\cite{wu2020tree} designed a tree-structured policy based RL model, where the root policy and leaf policy represent the coarse and fine decision-making steps respectively, to progressively regulate the predicted moment locations.
% \noindent\textbf{\textit{RL-based Methods.}} Some recent models regard the TSGV task as a sequential decision making problem, and resort to Reinforcement Learning (RL) algorithms. Specifically, Wang~\etal~\cite{wang2019language} introduced a semantic matching RL (SM-RL) model by extracting semantic concepts of videos and fusing them with global context features. Then, video contents are selectively observed and associated with the given sentence in a matching-based manner.  Hahn~\etal~\cite{hahn2019tripping} presented TripNet, which uses RL to efficiently localize relevant activity clips in long videos, by learning how to intelligently hop around the video. Wu~\etal~\cite{wu2020tree} formulated a tree-structured policy based progressive RL (TSP-PRL) model to sequentially regulate the predicated temporal boundaries by an iterative refinement process.

\textbf{\textit{Weakly Supervised Methods.}} 
Since the annotation process for temporal boundaries of retrieved moments is labor-intensive and costly, some studies resort to address the TSGV problem with only the video-level descriptions available for training~\cite{duan2018weakly,gao2019wslln,mithun2019weakly,song2020weakly,tan2019wman}. This kind of setting is dubbed as weakly supervised TSGV.
TGA~\cite{mithun2019weakly} learns a joint embedding network to align the text and video features, where the global visual features are obtained by weighted pooling according to the text-guided attentions. Duan~\etal~\cite{duan2018weakly} established a cycle system that consists of the weakly supervised localization task and its dual problem (\ie, weakly supervised dense event captioning) and minimized the reconstruction error for training such a loop system. Huang~\etal~\cite{huang2021cross} presented a cross-sentence relations mining~(CRM) method that explores the cross-sentence relations in the multi-sentence paragraph to improve the per-sentence grounding accuracy.
% Duan~\etal~\cite{duan2018weakly} formulate and address the problem of weakly supervised dense event captioning in videos~(\ie, to detect and describe all events of interest in a video), which is a dual problem of weakly supervised TSGV. It presents a cycle system to train the model which can solve such a pair of dual problems at the same time. In other words, weakly supervised TSGV can be regarded as an intermediate task in such a cycle system. 
% \noindent\textbf{\textit{Weakly Supervised Methods.}} Since the ground-truth annotations for the TSGV task are manually consuming, some works start to extend this problem to a weakly supervised scenario where the ground-truth segments are unavailable in the training stage~\cite{duan2018weakly,gao2019wslln,mithun2019weakly,song2020weakly,tan2019wman}. Mithun~\etal~\cite{mithun2019weakly} utilized a latent alignment between video frames and sentence descriptions with Text-Guided Attention (TGA), and TGA was used during the test stage to retrieve relevant moments. Duan~\etal~\cite{duan2018weakly} took the TSGV task as an intermediate step for dense video captioning, and then they established a cycle system and leveraged the captioning loss to train the whole model. Song~\etal~\cite{song2020weakly} presented a multi-level attentional reconstruction network, which leverages both intra- and inter-proposal interactions to learn a language-driven attention map, and can directly rank the candidate proposals at the inference stage.

\subsection{Biases in Temporal Sentence Grounding in Videos}
Recently, there are many works that are related to uncovering some forms of biases in TSGV datasets~\cite{otani2020uncovering,zhou2021embracing,nan2021interventional,yang2021deconfounded}. Otani~\etal~\cite{otani2020uncovering} revealed that the mainstream datasets have latent biases on ground-truth moment locations and current deep models are good at making use of them. Yang~\etal~\cite{yang2021deconfounded} stated that it is the location variable that causes the spurious correlation between video moments and predicted scores as a confounder, and they further presented a deconfounded cross-modal matching network to remove the confounding effects of the moment location. However, these works either just point out the problem of dataset biases in TSGV without a solution or design a debiased model without careful thinking about current evaluation protocols.

Moreover, Zhou~\etal~\cite{zhou2021embracing} were devoted to dealing with another kind of bias, \ie, the single-style of annotations. The proposed DeNet with a debiasing mechanism can produce diverse yet plausible predictions. Nan~\etal~\cite{nan2021interventional} proposed an approach to approximate the latent confounder set distribution based on the theory of causal inference to deconfound selection biases introduced by datasets (\eg, in datasets, it appears more often that a person is holding a vacuum cleaner than a person is repairing a vacuum cleaner). However, these two works~\cite{zhou2021embracing,nan2021interventional} can not resolve the issue of moment annotation distribution biases in TSGV.

Different from the above relevant studies, we not only raise the location bias issue and design new evaluation protocols including re-organized datasets and more reliable metrics, but propose a new debiasing framework from the perspective of causality to resolve the problem as well.
% The works above either                                                                                                                                                                                                                                                                                                                                                                                                    
% \textcolor{red}{balance the space of 2.2 and 2.3}

% \textcolor{blue}{Besides, there are also two works~\cite{nan2021interventional, yang2021deconfounded} introducing causal intervention in debiasing TSGV domain.} The former one~\cite{nan2021interventional} focuses on the interaction bias between two entities (\eg, it appears more often that a person is holding a vacuum cleaner than a person is repairing a vacuum cleaner), and the latter one ~\cite{yang2021deconfounded} is most closely related to our debiasing work, but with certain shortcomings, such as their relatively less well-designed out-of-distribution test, which will be subsequently compared in more detail.

\subsection{Biases in Other Tasks}
Besides TSGV, the dataset bias issue has been observed and addressed in many other multimedia tasks~\cite{cadene2019rubi,clark2019don,agrawal2018don,grand2019adversarial,niu2021counterfactual,yang2021deconfounded-image, tang2020unbiased}.
% to improve the robustness of models.

In Visual Question Answering~(VQA), due to the unbalanced distribution of answers, some models are able to give fairly good answers without understanding the visual contents. Thus, a new data split namely VQA-CP (under Changing Priors)~\cite{agrawal2018don} that alters the language prior distribution is proposed to evaluate the generalization ability of models. In VQA-CP dataset, the answer distribution for each question type in the test set is different from that in the training set. To avoid exploiting the language biases, some ensemble-based methods including fusion-based approaches~\cite{gokhale2020mutant,cadene2019rubi,clark2019don,agrawal2018don,chen2020counterfactual,chen2021counterfactual} and adversarial-based approaches~\cite{grand2019adversarial, ramakrishnan2018overcoming} have emerged. 

However, these debiased VQA methods are not able to give a formal formulation of the bias. CF-VQA~\cite{niu2021counterfactual} creatively revisits the methods above from a causal perspective, formulating the language biases as the direct causal effect of questions on answers, and it further presents a novel counterfactual inference framework. The causality can provide good interpretability and theoretical support for debiasing strategies. Such causality-based debiasing idea has also inspired other fields~\cite{tang2020unbiased,yang2021deconfounded-image}. Tang~\etal~\cite{tang2020unbiased} proposed an unbiased method from biased training for Scene Graph Generation (SGG). Specifically, after analyzing the causal graph, they attempt to remove the harmful bias by computing Natural Direct Effect with counterfactual causality. Yang~\etal~\cite{yang2021deconfounded-image} analyzed the hidden cause in image captioning and pointed out the confounder is the pre-training dataset. They further presented DICv1.0 framework with both front-door and back-door adjustment.

\section{Revisiting Evaluation Protocols}
In this section, we perform a deep analysis on the limitations of current evaluation protocols including the benchmark datasets and metrics in Section~\ref{sec:dataset_analysis}. To address such limitations, we propose new and more trustworthy evaluation protocols in Section~\ref{sec:new-eval}.
\subsection{Analysis of Current Evaluation Protocols}
\label{sec:dataset_analysis}
To figure out where the specific biases come from and why the metrics cause unreliable model evaluation, we thoroughly analyze the datasets and metrics that are commonly adopted in TSGV.

\begin{figure}[!t]
   \begin{minipage}{0.65\textwidth}
     \centering
     \includegraphics[width=.98\linewidth]{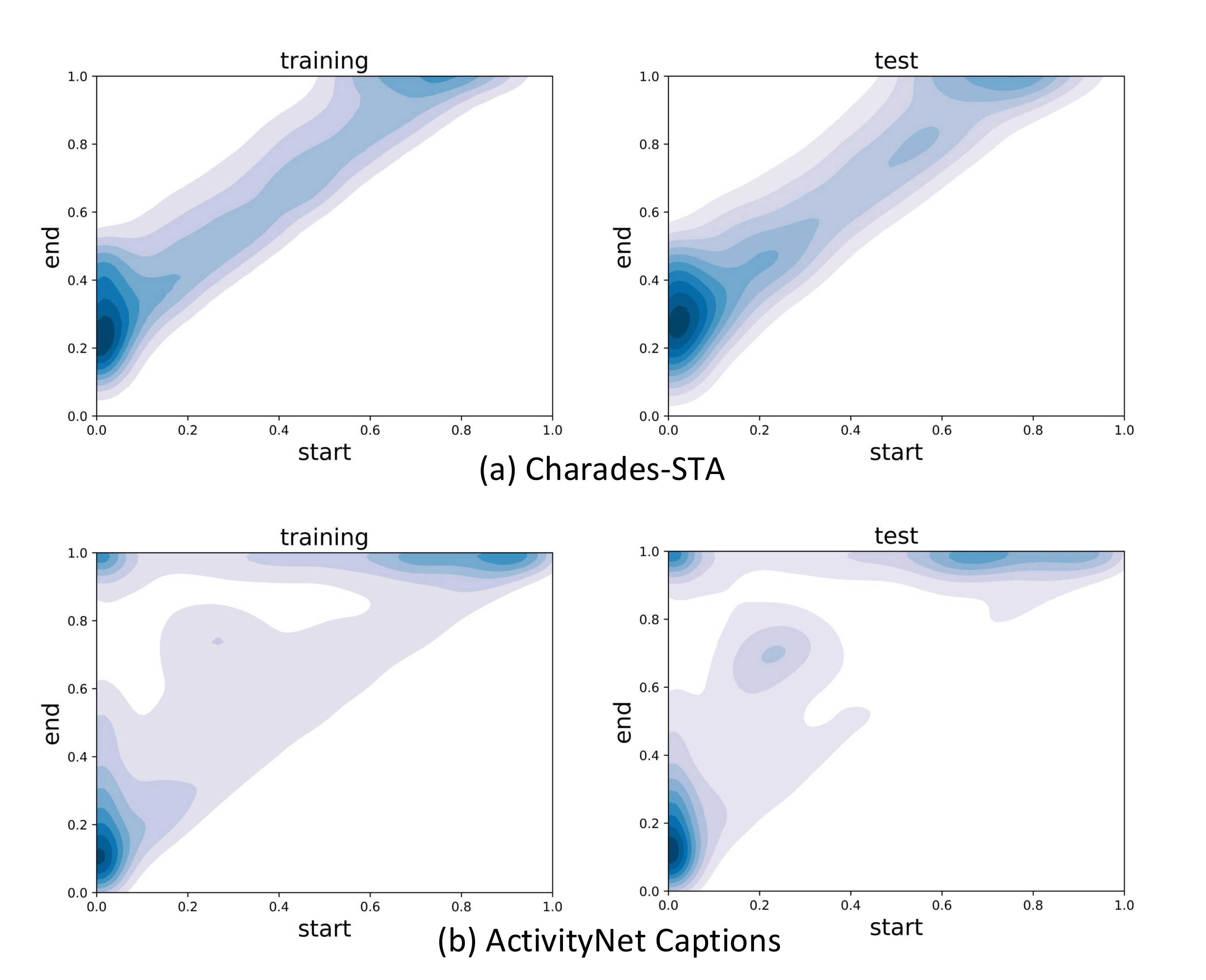}
     \caption{The ground-truth moment annotation distributions of all query-moment pairs in Charades-STA and ActivityNet Captions. The deeper the color, the larger density in distributions.}\label{fig:origin_split_dataset}
   \end{minipage}\hfill
   \begin{minipage}{0.32\textwidth}
     \centering
     \includegraphics[width=.95\linewidth]{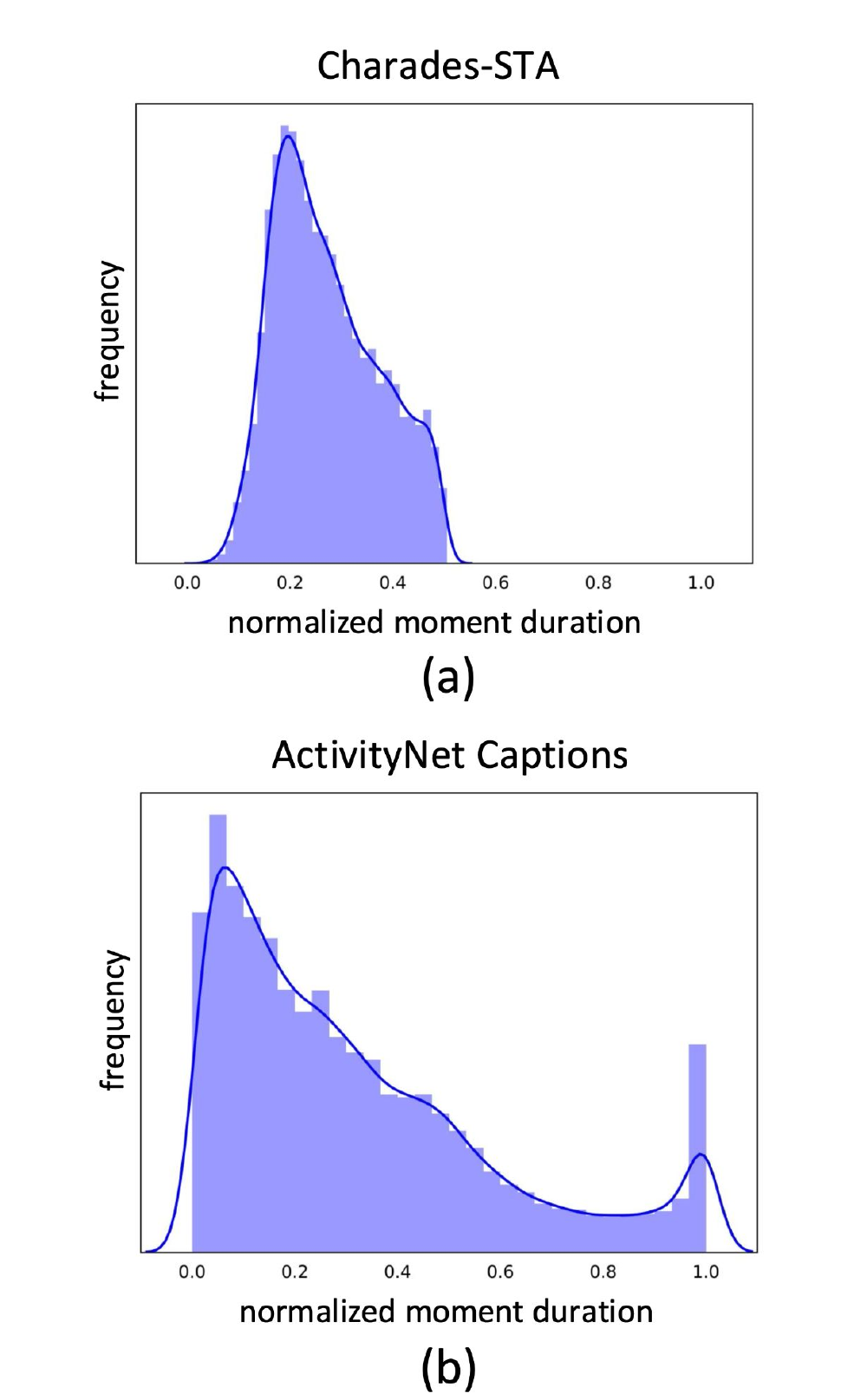}
     \caption{The histogram of the normalized ground-truth moment durations in Charades-STA and ActivityNet Captions.}
     \label{fig:duration}
   \end{minipage}
\end{figure}

\subsubsection{Datasets}
In TSGV research communities, four public datasets are widely used for evaluation, \ie, \textbf{TACoS}~\cite{regneri2013grounding}, \textbf{DiDeMo}~\cite{anne2017localizing},  \textbf{Charades-STA}~\cite{gao2017tall} and \textbf{ActivityNet Captions}~\cite{krishna2017dense}. However, some of them have obvious and inherent shortcomings, \eg, the video scene is restricted into the kitchen domain in TACoS dataset, and the ground-truth moments are comprised of the five-second video segment units in DiDeMo dataset. Therefore, the remaining two datasets (\ie, Charades-STA and ActivityNet Captions) have become the mainstream datasets for TSGV evaluation~\cite{chen2018temporally,hahn2019tripping,xu2019multilevel,yuan2019semantic,zeng2020dense,zhang2020learning}, which are also what we focus on.

% TODO: splits are of/  cover various complex human activities
\begin{wraptable}{r}{0.5\textwidth}
\raggedright
	\caption[]{The detailed statistics of datasets, including the number of videos and query-moment pairs for each data split.}
	\label{tab:dataset}
		\scalebox{0.85}{
			\begin{tabular}{l|lrr}
                \hline
                    Dataset & Split & \# Videos & \# Pairs  \\
                \hline
                    \multirow{2}{*}{Charades-STA} & training    & 5,338     & 12,408  \\
                    & test & 1,334 & 3,720 \\
                \hline
                    \multirow{2}{*}{ActivityNet Captions} & training & 10,009 & 37,421  \\
                    & test & 4,917 & 34,536 \\
                \hline
                    \multirow{4}{*}{Charades-CD (Ours)} & training & 4,564 & 11,071  \\
                    & val & 333 & 859 \\
                    & test-iid & 333 & 823 \\
                    & test-ood & 1,442 & 3,375 \\
                \hline
                    \multirow{4}{*}{ActivityNet-CD (Ours)}  & training  & 10,984    & 51,415 \\
                    & val & 746 & 3,521 \\
                    & test-iid & 746 & 3,443 \\
                    & test-ood & 2,450 & 13,578 \\
                \hline
			\end{tabular}
		}
\end{wraptable}
Below are more details about Charades-STA and ActivityNet Captions datasets. \textbf{Charades-STA}~\cite{gao2017tall} is built upon the original Charades dataset~\cite{sigurdsson2016hollywood}, focusing on those videos containing indoor daily activities. Its video length is around 30 seconds on average. The training/test splits are of 12,408/3,720 query-moment pairs. \textbf{ActivityNet Captions}~\cite{krishna2017dense} is extended from ActivityNet v1.3 dataset~\cite{caba2015activitynet} for dense event captioning. The videos cover various complex human activities. Each video is annotated with multiple descriptive sentences and their corresponding temporal boundaries of video moments.  Since the test split is withheld for the public competition challenge, the two accessible validation sets (\ie, ``val 1'', ``val 2'') are commonly merged as a test set for the TSGV evaluation. The training/test splits are of 37,421/34,536 query-moment pairs, respectively. 
% The videos in Charades-STA are 30-second long on average. In total, there are 12,408 and 3,720 query-segment pairs for training and test, respectively. \textbf{AcitivtyNet Captions}~\cite{krishna2017dense} is built on top of ActivityNet v1.3 dataset~\cite{caba2015activitynet}. The videos in this dataset cover a wide range of complex human activities. For each video, the temporal segment and caption sentence of each human event is annotated. Since the test split is withheld for competition, we follow the previous methods and merge the two validation subsets ``val 1'', ``val 2'' as the test set. The numbers of query-segment pairs for training and test split are 37,421 and 34,536, respectively.

% cha
% , which is extended from Charades~\cite{sigurdsson2016hollywood} by Gao~\etal~\cite{gao2017tall} for TSGV task, with hundreds of people recording videos in their own homes, acting out casual everyday activities, most of the moments start at very beginning of the videos and end at around $20\%-40\%$ of the length of the videos.

% anet 
% which is originally developed for dense video captioning~\cite{krishna2017dense}, covering a wide range of complex human activities, and each video is accompanied with a series of temporally annotated sentences, describing events that occur,

% \textcolor{red}{(Reorganize: simply introduce the datasets and then analysis the distribution.)} 
% TODO: combine with current description.

We visualize the joint distribution of normalized start and end points of the ground-truth moments (\cf.~Fig.~\ref{fig:origin_split_dataset}) in both datasets. An obvious observation is that the distributions of training and test sets for each dataset are almost the same, in other words, these two sets follow the independent and identical distribution (iid). We can also observe that each dataset has its own characteristics of the biased distribution. For Charades-STA, as we can see from Fig.~\ref{fig:origin_split_dataset} (a) that the vast majority of ground-truth moments are shorter than 0.5 (after normalization). The fact that the high-density parts concentrate on top-right and bottom-left corners indicates that moments are likely to be either at the beginning/end of the whole videos. For ActivityNet Captions (\cf, Fig.~\ref{fig:origin_split_dataset} (b)), there are mainly three types of ground-truth moments appearing more frequently: Short moment samples ($\leq 0.3$ after normalization) that start either at the beginning or end of the videos and overlong moment samples that nearly cover the whole length (top-left corner). The main reason for so many overlong samples in ActivityNet Captions is that this dataset is originally created for dense video captioning, which should be annotated with video-level captions. Table~\ref{tab:dataset} shows more detailed statistics about these two datasets. In summary, both of these two datasets have strong biases of the ground-truth moment distribution. A simple baseline method that only exploits such biases may be able to achieve competitive results with SOTA models (\cf, Fig.~\ref{fig:intro}).

\subsubsection{Evaluation Metrics} \label{old_metric}
% half re-write
The commonly used evaluation metric for assessing the moment localization results in TSGV is ``R@$n$,IoU@$m$". It measures the percentage of positive samples out of all testing samples, which is formally defined as:
\begin{equation}
\text{R@$n$,IoU@$m$} = \frac{1}{N_q} \sum_{i} r(n,m,q_i) \,,
\end{equation}
where for each query $q_i$, $r(n,m,q_i)=1$ if at least one of the top-$n$ predicted moments has an IoU (Intersection-over-Union) larger than threshold $m$ with the ground-truth moment, otherwise $r(n,m,q_i)=0$. The total number of all samples is $N_q$.

% . Specifically, , it first computes the temporal Intersection-over-Union (IoU) between the predicted moment and its ground-truth, and this metric

% half re-write
Some existing works~\cite{chen2018temporally,liu2018cross,xu2019multilevel,yuan2019find,zhang2020learning} report the metric scores with some small IoU thresholds like $m \in \{0.1,0.3,0.5\}$. However, such metrics with small IoU thresholds may overrate the model performance when datasets have obvious annotation biases. As shown in Fig.~\ref{fig:duration}.~(b), for ActivityNet Captions, a substantial proportion of ground-truth moments occupy a long period of video duration. In statistics, 40\%, 20\%, and 10\% of queries refer to a moment occupying over 30\%, 50\%, and 70\% duration of the entire video, respectively. Such annotation biases can increase the chance of hitting the ground-truth moments when IoU thresholds are small. Taking an extreme case as an example, when the IoU threshold is 0.3, if the ground-truth moment is the entire video, any predictions with a duration longer than 0.3 can be seen as positive. Thus, the metric ``R@$n$,IoU@$m$'' with small $m$ is unreliable for current biased annotated datasets.
% Most of previous TSGV methods~\cite{chen2018temporally,liu2018cross,xu2019multilevel,yuan2019find,zhang2020learning} always report their scores on some small IoU thresholds like $m \in \{0.1,0.3,0.5\}$. However, as shown in Figure~\ref{fig:duration} (b), for dataset ActivityNet Captions, a substantial proportion of ground-truth moments have relatively long durations. Statistically, 40\%, 20\%, and 10\% of sentence queries refer to a moment occupying over 30\%, 50\%, and 70\% of the length of the whole video, respectively. Such annotation biases can obviously increase the chance of correct predictions under small IoU thresholds. Taking an extreme case as example, if the ground-truth moment is the whole video, any predictions with duration longer than 0.3 can achieve R@$1$,IoU@$0.3 = 1$. Thus, metric R@$n$,IoU@$m$ with small $m$ is unreliable for current biased annotated datasets.

\begin{figure*}[!t]
	\centering
	\includegraphics[width=0.9\textwidth]{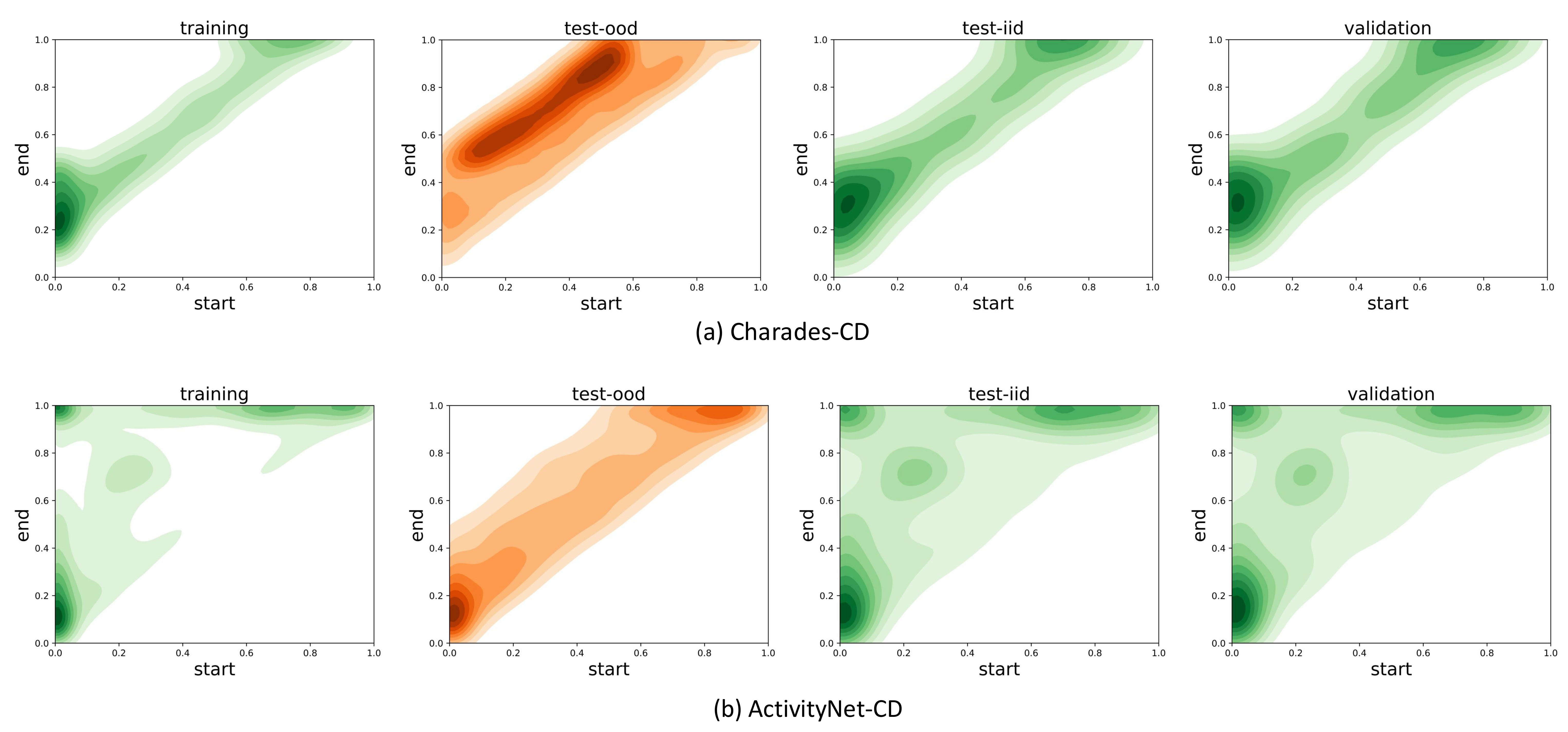}
	\caption{(a) and (b) illustrate the ground-truth moment annotation distributions of each split in two re-organized datasets.}
	\label{fig:resplit_dataset}
\end{figure*}

\subsection{New Evaluation Protocols} \label{sec:new-eval}
% half re-write
In order to overcome the shortcomings of current evaluation protocols, we come up with solutions for both the datasets and metrics. As for datasets with obvious annotation biases, we propose to re-organize them, deliberately changing the moment location distribution in the test set. As for unreliable evaluation metrics with small IoU thresholds, we design new metrics to rectify the overrating performance scores.

\subsubsection{Dataset Re-splitting}
We propose to re-organize the two datasets (\ie, Charades-STA and ActivityNet Captions), naming the re-organized ones as \textbf{Charades-CD} and \textbf{ActivityNet-CD} (CD means \textbf{C}hanging \textbf{D}istribution), respectively. To be specific, each dataset is re-split into four sets, \ie, \textbf{training},
\textbf{validation (val)}, \textbf{test-iid}, and \textbf{test-ood}. We make all samples from the training, val, and test-iid sets follow the independent and identical distribution, and make the samples of test-ood set out-of-distribution. Obviously, the performance gap between the test-iid set and test-ood set can effectively evaluate the generalization capability of the model. The following parts further describe the details during the process of data re-splitting.
% To accurately monitor the research progress in TSGV and reduce the influence of moment annotation biases, we propose to re-organize the two datasets (\ie, Charades-STA and ActivityNet Captions) by deliberately assigning different moment annotation distributions in each split. Particularly, each dataset is re-splitted into four sets: \textit{training}, \textit{validation (val)}, \textit{test-iid}, and \textit{test-ood}. All samples in the training, val, and test-iid sets satisfy the independent and identical distribution, and the samples in test-ood set are out-of-distribution. The performance gap between the test-iid set and test-ood set can effectively reflect the generalization ability of the models. We name the two new re-organized datasets as \textbf{Charades-CD} and \textbf{ActivityNet-CD}.

\textbf{Dataset Aggregation and Splitting.}
% half re-write
For each dataset, we collect all the query-moment pairs (samples) in the training and test sets, and use the Gaussian kernel density estimation to fit the moment annotation distribution as mentioned in Section~\ref{sec:dataset_analysis} (\cf, Fig.~\ref{fig:origin_split_dataset}). Afterwards, we sort all the samples based on their probabilistic density values (from high to low), and take the lowest 20\% samples as the preliminary test-ood set since the distribution is furthest different from that of the whole dataset. The remaining 80\% samples are divided into the preliminary training set.

\textbf{Conflicting Video Elimination.} 
Since each video is associated with several sentence queries (samples), it is necessary to ensure that no video simultaneously appears in both the training and test sets. Thus, after obtaining the preliminary test-ood set, we check whether the videos of test-ood samples are also in the preliminary training set. If so, we move all samples (\ie, query-moment pairs) referring to the same video into the split with most of samples. In addition, to avoid the inflating performance of overlong predictions in ActivityNet-CD (\cf, the \texttt{PredictAll} baseline in Fig.~\ref{fig:intro}), we leave all samples with ground-truth moment occupying over 50\% video duration in the training set.

After elimination of all conflicting videos, the final test-ood set occupying around 20\% query-moment pairs of the entire dataset is obtained. Then, we randomly split the remaining samples (based on videos) into three groups for the collection of the training, val, and test-iid sets, which occupy around 70\%, 5\%, and 5\% samples, respectively. More detailed statistics of the re-organized datasets should be found in Table~\ref{tab:dataset}.

% After eliminating all conflicting videos, we obtain the final test-ood set, which consists of around 20\% query-moment pairs of the whole dataset. Then, we randomly divide the remaining samples (based on videos) into three splits: the training, val, and test-iid sets, which consist of around 70\%, 5\%, and 5\% data samples, respectively. The statistics of the new proposed splits are reported in Table~\ref{tab:dataset}.
% After eliminating the joint videos, we obtain the final test-ood set, and we also randomly sample some videos as well as their query-segment pairs from the candidate training set to compose the validation and independent and identical test (test-iid) sets. Finally, we obtain the test-ood set containing about 20\% query-segment pairs of the overall samples, and other 70\%, 5\% and 5\% pairs are in the resplit training, test-iid, and validation sets, respectively. The statistics of the original and the new data splits are provided in Table~\ref{tab:dataset}.
% small changes here
\begin{figure*}[!t]
	\centering
	\includegraphics[width=0.9\columnwidth]{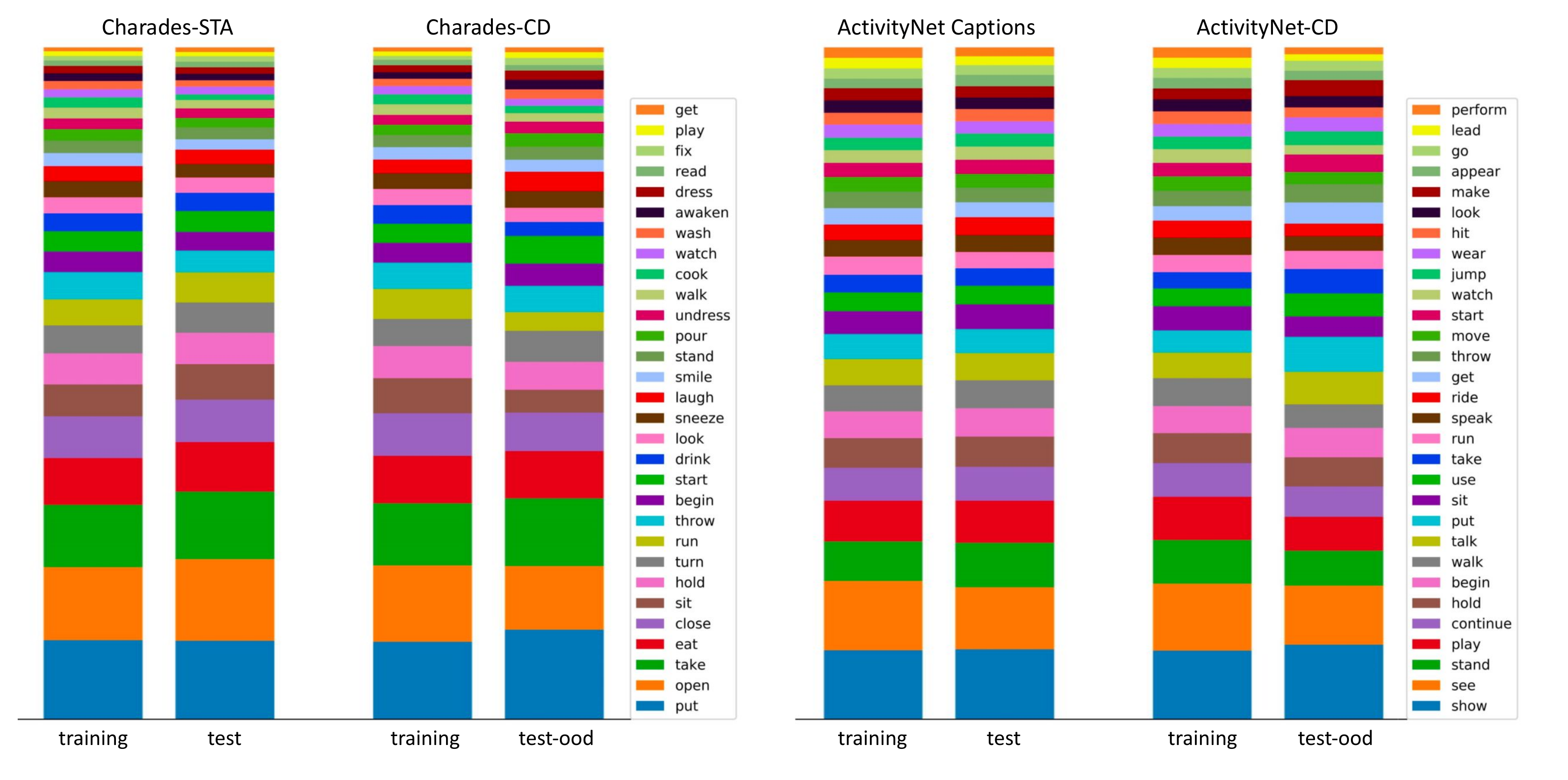}
	\caption{Top-30 frequent actions in training/test splits for each dataset. The longer the bar, the more frequently the action appears.}
	\label{fig:verb}
\end{figure*}

\begin{figure*}[!b]
	\centering
	\includegraphics[width=0.95\textwidth]{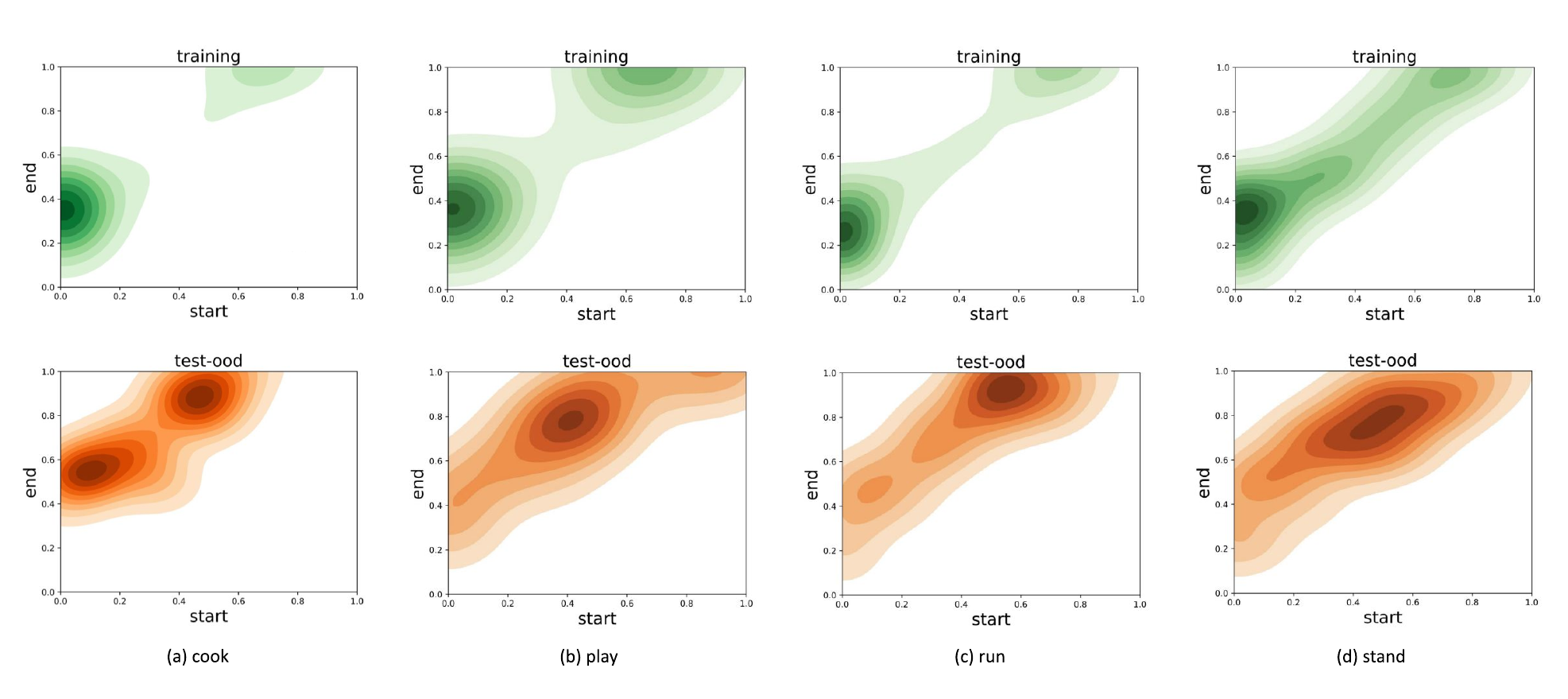}
	\caption{The moment annotation distributions of the query-indicated moments which contain a specific action (\eg, \textit{cook}) in the training and test-ood sets of Charades-CD. The deeper the color, the larger the density in the distribution.}
	\label{fig:resplit_dataset_action}
\end{figure*}

\textbf{New Split Analysis.} 
Fig.~\ref{fig:resplit_dataset} depicts the ground-truth moment distributions of these two re-organized datasets. An obvious observation is that the annotation distributions of test-ood set (best expressed in orange) are significantly different from others while the distributions of other three sets (best expressed in green) are similar with those of original training/test splits (\cf, Fig.~\ref{fig:origin_split_dataset}). We investigate the difference of the proposed test-ood split for each of these two datasets: 1) For Charades-CD, the start points of the ground-truth moments are distributed more diversely, instead of concentrating at the beginning of the videos. 2) For ActivityNet-CD, instead of concentrating in three corners, there are more samples locating in relatively central areas so that models will fail to perform well by merely exploiting the 
moment distribution biases.

We also investigate the action distribution in each of the original and re-organized datasets. We count the frequency of each verb occurring in the sentence queries of each split, which obviously forms a long-tail distribution. Then the top-30 frequent verbs are shown in Fig.~\ref{fig:verb}, with action coverage of 92.7\% and 52.9\% for Charades-CD and ActivityNet-CD, respectively. We can observe that the action distribution of new test-ood set is still similar with that of either original or re-organized training split for each datasets, which indicates that the OOD comes from each specific verb. As shown in Fig.~\ref{fig:resplit_dataset_action}, for a given verb, the moment annotations of the training and test-ood sets are of significantly different distributions.

\subsubsection{Proposed Evaluation Metric}
\begin{wrapfigure}{r}{0.5\textwidth}
	\centering
	\includegraphics[width=0.43\columnwidth]{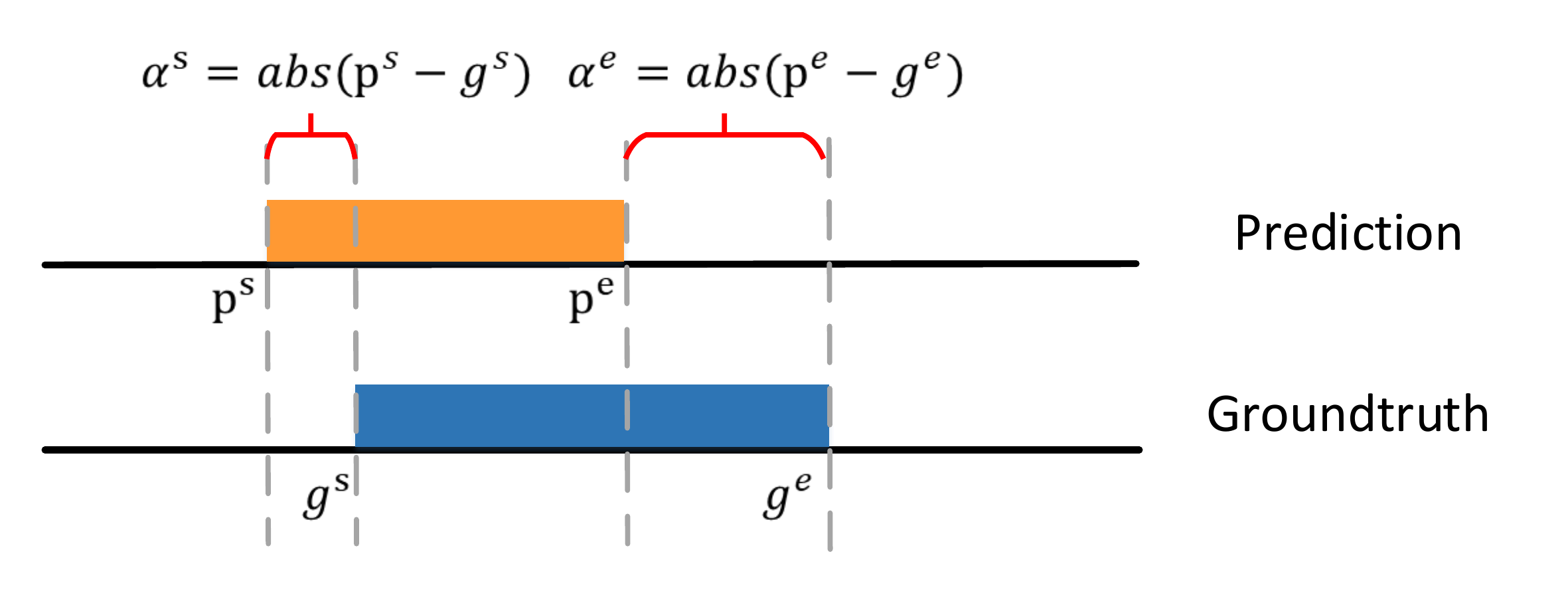}
	\caption{An illustration of the proposed ``dR@$n$,IoU@$m$'' metric.}
	\label{fig:newmetric}
\end{wrapfigure}

% It penalizes inaccurate coordinate predictions by adding two product terms ranging from 0 to 1. The terms measure the temporal distance of the start/end points between the predicted moment and gt moment, regularized by the video length.
As discussed in Section~\ref{old_metric}, the most prevailing evaluation metric --- R@$n$,IoU@$m$ --- is untrustworthy under small threshold $m$. To alleviate this issue, as shown in Fig.~\ref{fig:newmetric}, we propose to calibrate the $r(n,m,q_i)$ value by considering the ``temporal distance'' between the predicted and ground-truth moments. Specifically, we propose a new metric discounted-R@$n$,IoU@$m$, denoted as ``dR@$n$,IoU@$m$'':
\begin{equation}
\label{eq:new_metric}
\text{dR@$n$,IoU@$m$} = \frac{1}{N_q}  \sum_{i}r(n,m,q_i) \cdot \alpha^s_i \cdot \alpha^e_i, 
\end{equation}
where $\alpha^*_i = 1-\text{abs}(p_i^*-g_i^*)$, and $\text{abs}(p_i^*-g_i^*)$ is the absolute distance between the boundaries of predicted and ground-truth moments. Both $p_i^*$ and $g_i^*$ are normalized to the range (0, 1) by dividing the video duration. When the predicted and ground-truth moments are very close to each other, the discount ratio $\alpha^*_i$ will be close to 1, \ie, the new metric can degrade to ``R@$n$,IoU@$m$'' with exactly accurate predictions. Otherwise, even the IoU threshold condition is met, the score $r(n,m,q_i)$ will still be discounted by $\alpha^*_i$, which helps to alleviate the inflating recall scores under small IoU thresholds. With the proposed     ``dR@$n$,IoU@$m$'' metric, those speculation methods which over-rely on moments annotation biases (\eg, long moments annotations in ActivityNet Captions) will not perform well.

% Moment distributions of actions in different data splits of Charades-STA (Charades-CD)

\section{Proposed debiasing Approach}
\begin{figure*}[!t]
	\centering
	\includegraphics[width=1.0\textwidth]{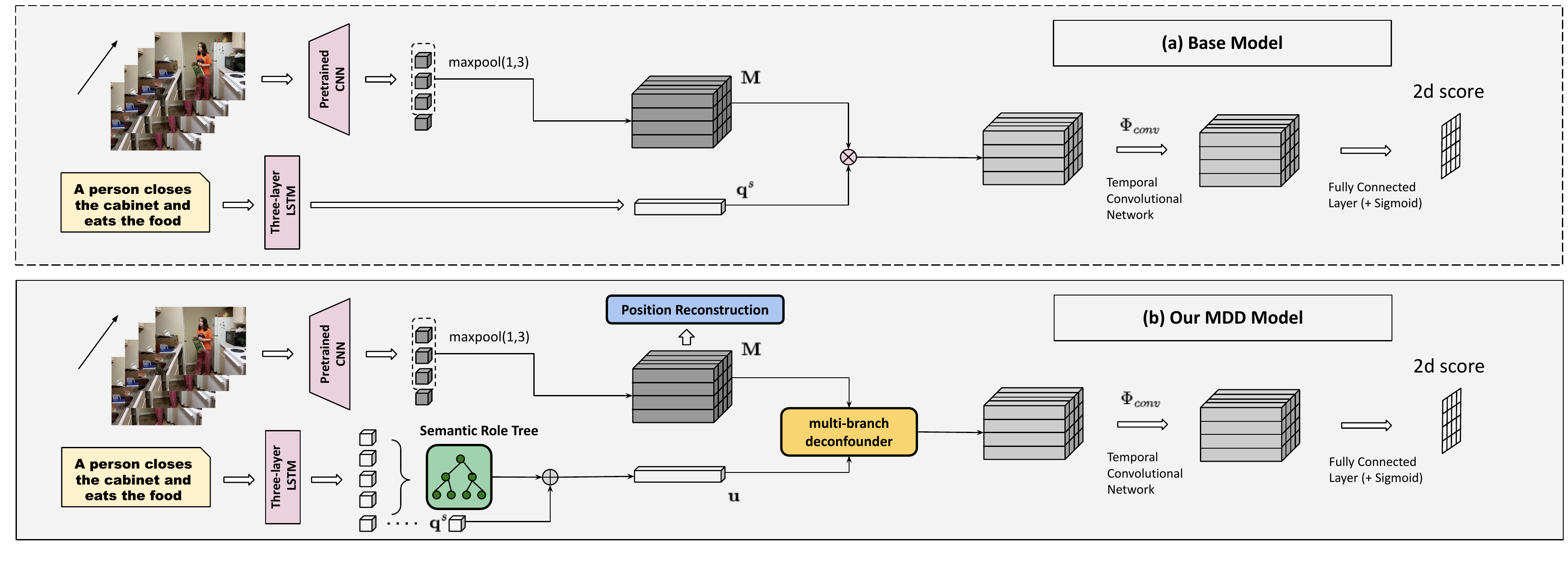}
 	\caption{The overall framework of the Multi-branch Deconfounding Debiasing (MDD) framework. Specifically, (a) briefly shows the pipeline of base 2D-TAN model~\cite{zhang2020learning}, based on which, we develop three important components indicated by ``semantic role tree'', ``position recontruction'' and ``multi-branch decondounder'', yielding our proposed MDD model as shown in (b). More specifically, (i) we enhance the moment representations with position reconstruction module and (ii) parse the query into the semantic role tree to get more fine-grained textual features. (iii) During the multimodal fusion process, we adopt a multi-branch deconfounder to remove the effects caused by multiple confounders.}
	\label{fig:framework}
\end{figure*}

To reduce the effects of moment annotation biases, we further propose a novel debiasing approach. The overall framework is shown in Fig.~\ref{fig:framework}. Basically, we add three key components to the base model for unbiased moment predictions. In this section, we firstly define the TSGV problem and illustrate how the base model works. Afterwards, each of the key components will be described in detail, along with ultimate learning objectives.

\subsection{Problem Formulation}
% fix the symbol definition, consistent with the following equations.
% \textcolor{red}{(Delete the feature extraction, and put it into the feature extraction part. Make sure each subsection/subfigure/paragraph is self-contained!!!It is very important!!!)} 
As shown in the example of Fig.~\ref{fig:intro}, a formal TSGV task takes a sentence query and an untrimmed video as inputs. The untrimmed video can be divided into multiple candidate moments. We let $Q$ denote the sentence query and $V$ denote the candidate video moments. For a proposal-based method which outputs the matching scores between the sentence query and each of candidate moments, a function $\mathcal{F}(Q,V)$ should be learned. The highest output score of the function indicates the best matching query-moment pair.
% Given the query, we first adopt pre-trained GloVe~\cite{pennington2014glove} to obtain corresponding word embeddings $\{s_1,s_2,\dots,s_{l_q}\}$. For the untrimmed video, we use pre-trained CNN model to extract features and get the moment-level features with pooling strategy as \cite{zhang2020learning} do. The moment features can be represented as $\{m_1, m_2,\dots,m_{l_v}\}$. All we need is to train a model to obtain the matching scores of query-moment pairs.

%  Firstly, the video clip features are extracted from a pre-trained CNN and form a 2D moment feature map via max pooling strategy, and the query features are obtained by feeding the GloVe word embeddings into a three-layer LSTM.

% Then the generated fused 2D temporal feature map is fed into the temporal convolutional network followed by fully-connected layer to obtain the final 2D matching scores.

\subsection{Base Model}
Due to the superior performance of 2D-TAN~\cite{zhang2020learning} in recent public models, we adopt it as the base model for our unbiased temporal sentence grounding. The core idea of 2D-TAN is utilizing a 2D feature map to represent candidate moments of various lengths and locations, where one dimension depicts the start indices of moments and the other one represents the end. 

More specifically, as shown in Fig.~\ref{fig:framework} (a), for the sentence query, it first embeds the words within the sentence query $S$ via GloVe~\cite{pennington2014glove} to obtain the corresponding word vectors, and then the word vectors are fed into a three-layer LSTM, where the last hidden state denoted as $\mathbf{q}^s \in \mathbb{R}^{d^h}$ is used to encode the whole query. For the video sequence, it first segments the video into non-overlapping clips, then samples the clips to a fixed size. The features of sampled $N^v$ video clips are extracted by a pre-trained CNN model and projected into the dimension of $d^v$, which can be denoted as $\{\mathbf{c}_1,\mathbf{c}_2,\dots,\mathbf{c}_{N^v}\}$. The moment feature $\mathbf{m}_{ij}$ ($1 \leq i \leq j \leq N^v$) out of the 2D feature map $\mathbf{M} \in \mathbb{R}^{N^v \times N^v \times d^v}$ can be obtained by adopting max pooling strategy on clips $\{\mathbf{c}_i,\mathbf{c}_{i+1},\dots,\mathbf{c}_{j}\}$. Afterwards, the 2D feature map $\mathbf{M}$ is fused with the query feature $\mathbf{q}^s$ and fed into a temporal adjacent network to model the temporal relations of moments. Then it passes through a fully connected layer and a Sigmoid function to generate the final 2D matching score map. 
% The temporal adjacent network aims to model the temporal relations of moments which is implemented by multi-layer convolutional network. fusion: hadmard product.

However, the inherent structure of 2D-TAN has natural advantages in exploiting location bias of datasets, since 2D feature map $\mathbf{M}$ is indexed by moment locations. Therefore, we propose to improve this base model from two aspects. On the one hand, due to the difficulties of semantic alignment between two modalities, the representation capability from each single modality should be enhanced. On the other hand, we attempt to debias the model from perspective of causality as causality-based methods have proven to be successful in debiasing from other fields. %cite?

\subsection{Improvements on Feature Extraction} \label{feature-extraction}
In order to improve the representation of both modalities, we propose to perform more detailed and effective feature extraction. Section~\ref{sec:fine-grained} depicts the process of extracting more fine-grained query feature, instead of directly involving the global query feature $\textbf{q}^s$ into subsequent multimodal fusion, a more fine-grained one denoted as $\textbf{q}^u$ is obtained based on $\textbf{q}^s$ and the sentence structure. Section~\ref{sec:position-reconstruct} illustrates how to enhance the 2D moment representations $\textbf{M}$ via position reconstruction, which aims to make it explicitly associated with the location attribute. 
% We replace the global query feature $\textbf{q}^s$ with gated fine-grained $\textbf{q}^u$, which will be depicted in detail in Section~\ref{sec:fine-grained}.
% TODO(shown in figure)

\subsubsection{Gated Fine-grained Query Feature} \label{sec:fine-grained}
\begin{figure*}[!t]
    \centering
    \includegraphics[width=.95\textwidth]{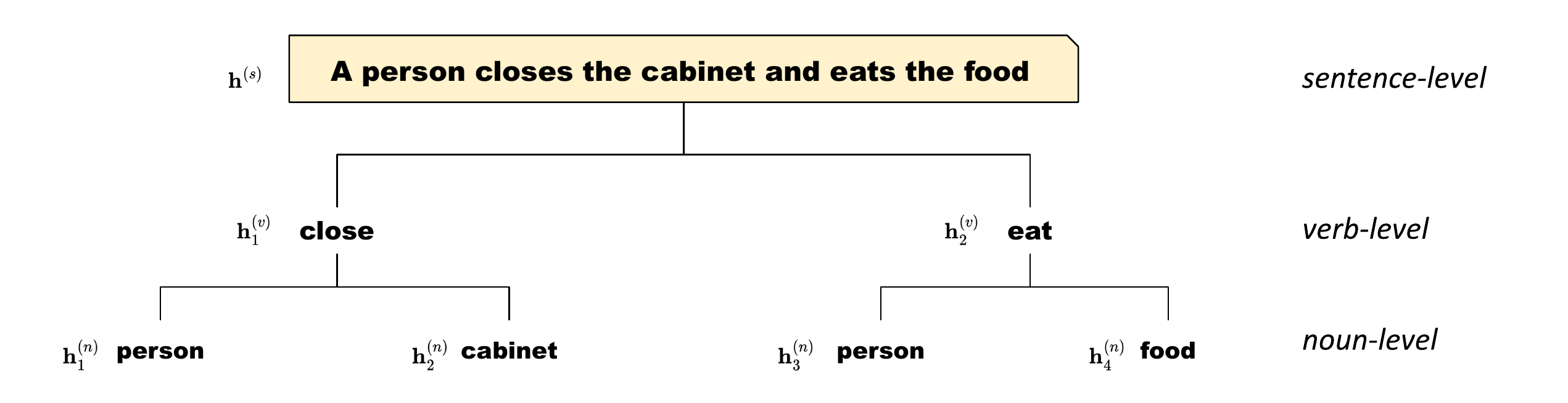}
    \caption{The parsed three-layer semantic role tree, where the root represents the whole sentence, and each subtree below the root represents a phrase centered by a verb with relevant objects (nouns) as leaf nodes.}
    \label{fig:tree}
\end{figure*}
% Diving Into The Relations: Leveraging Semantic and Visual Structures For Video Moment Retrieval & fast video moment retrieval (two papers)
In order to ultimately obtain the fine-grained query feature $\textbf{q}^u$, one off-the-shelf toolkit~\cite{gao2021fast,wu2021diving} is used to parse the sentence into a semantic role tree. By adopting hierarchical attention mechanism on the tree, we can get the phrase-level features $\{\textbf{g}_k\}_1^{N_{verb}}$. Then the phrase-level features are aggregated to obtain the final fine-grained sentence representation.
% As shown in Fig.~\ref{fig:framework}(c), we parse the whole sentence into a three-layer semantic role tree, in which the root node represents the whole sentence, and each subtree below it uses the predicate~(verb) as the root and the relevant objects~(nouns) as the leaf nodes.

\textbf{Attention from Sentence-level to Verb-level.} More specifically, we first initiate the representation $\mathbf{h} \in \mathbb{R}^{d^h}$ of each node with sequential outputs of the three-layer LSTM. Then we obtain the attention weights of verbs according to the root:
% \mathrm{Att}(\mathbf{h}^s, \mathbf{h}_i^v) \mathrm(Att)(\mathbf{k}, \mathbf{v}) =\\
\begin{equation}
\label{eq:tree_att}
    \gamma_k^{(v)} = \mathbf{W}_{\gamma}(\textrm{tanh}(\mathbf{W}_{top}\mathbf{h}^{(s)}||\mathbf{W}_{down}\mathbf{h}_k^{(v)})), \qquad k = 1,\dots,N_{verb} \,,
\end{equation}
where $\mathbf{W}_{top}\in \mathbb{R}^{d^h \times d^h}$, $\mathbf{W}_{down}\in \mathbb{R}^{d^h \times d^h}$ and $\mathbf{W}_{\gamma}\in \mathbb{R}^{1 \times 2d^h}$ are learnable variables. $\mathbf{h}^{(s)} \in \mathbb{R}^{d^h}$ is the global feature of the whole sentence (\ie, $\textbf{q}^s$) and $\mathbf{h}_k^{(v)} \in \mathbb{R}^{d^h}$ denotes the feature of the $k$-th verb node. $||$ implies the concatenation operation.

\textbf{Attention from Verb-level to Noun-level.} Then we aggregate all the verb nodes to obtain the global verb representation $\widetilde{\mathbf{h}}^{(v)} \in \mathbb{R}^{d^h}$ as:
\begin{equation}
\begin{split}
    \widetilde{\mathbf{h}}^{(v)} &= \sum_{k=1}^{N_{verb}}\alpha_k^{(v)}\mathbf{h}_k^{(v)} \,, \\
    \bm{\alpha^{(v)}} &= \textrm{Softmax}(\bm{\gamma^{(v)}}) \,.
\end{split}
\end{equation}

Afterwards, we use a similar attention module to obtain the attention weights of noun~(leaf) nodes:
\begin{equation}
    \gamma_l^{(n)} = \mathbf{W}_{\gamma}(\tanh(\mathbf{W}_{top}\widetilde{\mathbf{h}}^{(v)}||\mathbf{W}_{down}\mathbf{h}_l^{(n)})), \qquad l = 1,\dots,N_{noun} \,,
\end{equation}
where $\mathbf{h}_l^{(n)}$ denotes the feature of the $l$-th noun node. It is worth noting that $\mathbf{W}_{\gamma}$, $\mathbf{W}_{top}$ and $\mathbf{W}_{down}$ are the sharing parameters with Equation~\eqref{eq:tree_att}.

\textbf{Phrase-level Features.} Then the phrase-level representation of each subtree $\textbf{g}_k$ can be yielded by aggregating all nodes within the subtree based on the weights:
\begin{equation}
    \begin{split}
        \bm{\beta} &= \textrm{Softmax}(\gamma_k^{(v)},\gamma_{z_{k,1}}^{(n)},\gamma_{z_{k,2}}^{(n)},\dots, \gamma_{z_{k,{nv}_k}}^{(n)}) \,, \\
        \textbf{g}_k &= \beta_0 \mathbf{h}_k^{(v)} + \sum_{j=1}^{{nv}_k} \beta_j \mathbf{h}_j^{(n)} \,,
    \end{split}
\end{equation}
where $z_{k,*}$ is a set of indices to enumerate all leaf nodes of subtree $k$. 

Then all the subtree representations are aggregated to obtain the gating signal $\mathbf{\bar{g}}$, and finally the fine-grained sentence feature representation $\textbf{q}^u$ is obtained as follows:
\begin{equation}
    \begin{split}
        \bar{\textbf{g}} &= \frac{1}{N_{verb}}\sum_{k=1}^{N_{verb}}\textbf{g}_k\,, \\
        \textbf{q}^u &= \textbf{q}^s + \textbf{q}^s \odot \bar{\textbf{g}}\,.
    \end{split}
\end{equation}

\subsubsection{Enhanced Moment Representation via Position Reconstruction} \label{sec:position-reconstruct}
For visual information, in order to better discriminate video moments with unique position information, we attempt to decouple the positional feature from the video moment feature to enhance the moment representation. Specifically, we feed the 2D temporal moment feature map $\mathbf{M} \in \mathbb{R}^{N^v \times N^v \times d^v}$ into a fully-connected layer to obtain the learned 2D position embedding $\mathbf{M}_p \in \mathbb{R}^{N^v \times N^v \times d^p}$, and then we establish a reconstruction loss function to make $\mathbf{M}_p$ close to the 2D position encoding $\mathbf{M}_e \in \mathbb{R}^{N^v \times N^v \times d^p}$:
\begin{equation}
    \begin{split}
        \mathbf{M}_p &= \tanh(\textrm{FC}(\mathbf{M}))\,, \\
        \mathcal{L}_{recon} &= ||\mathbf{M}_p - \mathbf{M}_e||_2 .
    \end{split}
\end{equation}
Here, $||\cdot||_2$ denotes L2-norm, 2D position encoding $\mathbf{M}_e$ is computed by sine and cosine functions of the different frequencies following \cite{vaswani2017attention} and $d^p$ denotes the dimension of positional features. 

% \noindent\textbf{Analysis on Single Confounder.}
% Inspired by the work~\cite{yang2021deconfounded}, we leverage the structured causal model to analyze the underlying relations among all variables of this TSGV problem. The causal graph which is a directed acyclic graph is shown in Fig.~\ref{fig:causal-graph}, where the nodes denote the variables and the directed edges denote the relations between nodes. $Q$ is the variable of query, $V$ denotes the video moment and $Y$ is the variable of predicted matching score. For those traditional TSGV models, they train a model to obtain the probabilities $P(Y|Q,V)$ that is conditioned on $Q$ and $V$. However, there may exist a confounder $C$ that has connections with both the multimodal inputs (\ie, $V$ and $Q$) and output scores $Y$. The confounder is harmful since it causes spurious correlation between the inputs and outputs. Thus, in order to remove the negative effects caused by the confounder, in the theory of causality, causal intervention is adopted with the do-calculus operation to block the effects:
% \begin{equation}
%     P(Y|do(Q,V)) = \sum_{c}P(Y|Q,V,c) \cdot P(c) \,.
% \end{equation}

% TODO: one sentence to explain the equation.

% empirically set the size of 80
\begin{figure*}[!t]
\centering
\subfloat[]{
\includegraphics[height=4cm]{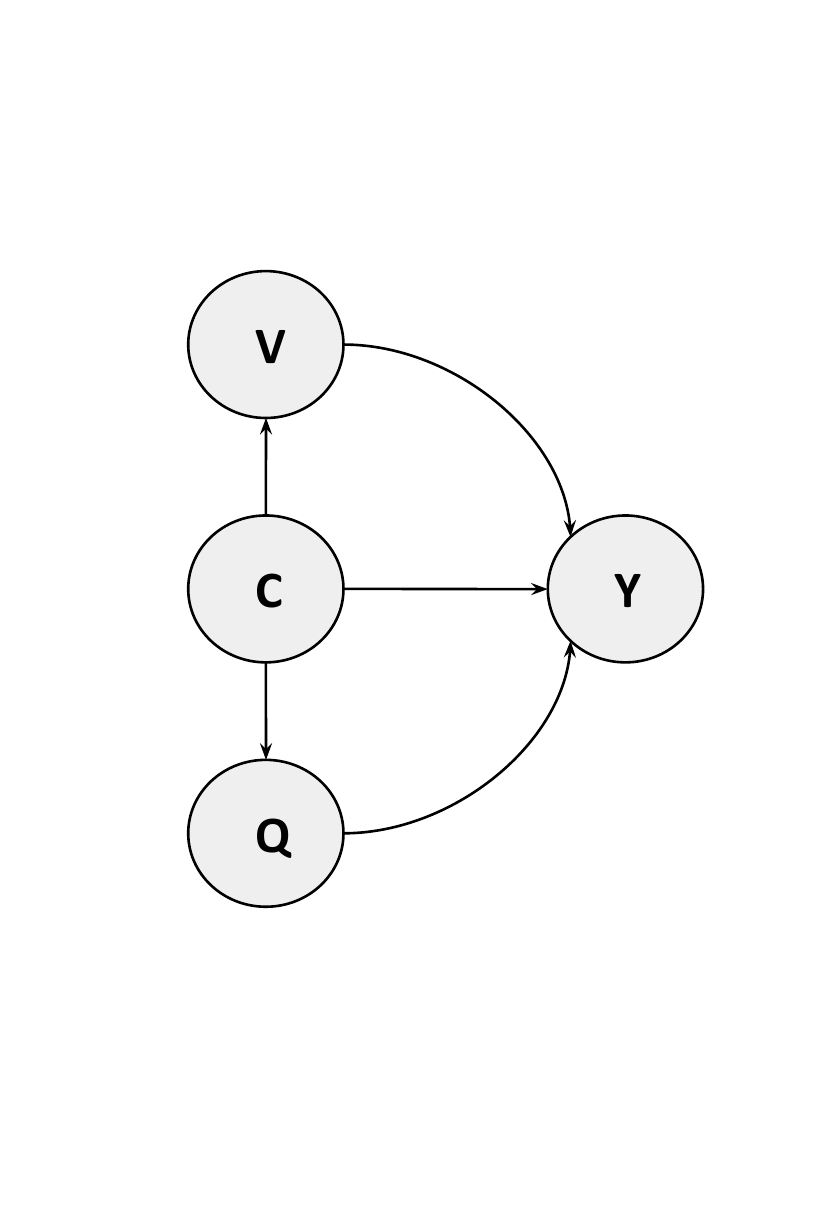}%
\label{fig:causal-graph}
}
\hfil
\subfloat[]{
\includegraphics[height=4cm]{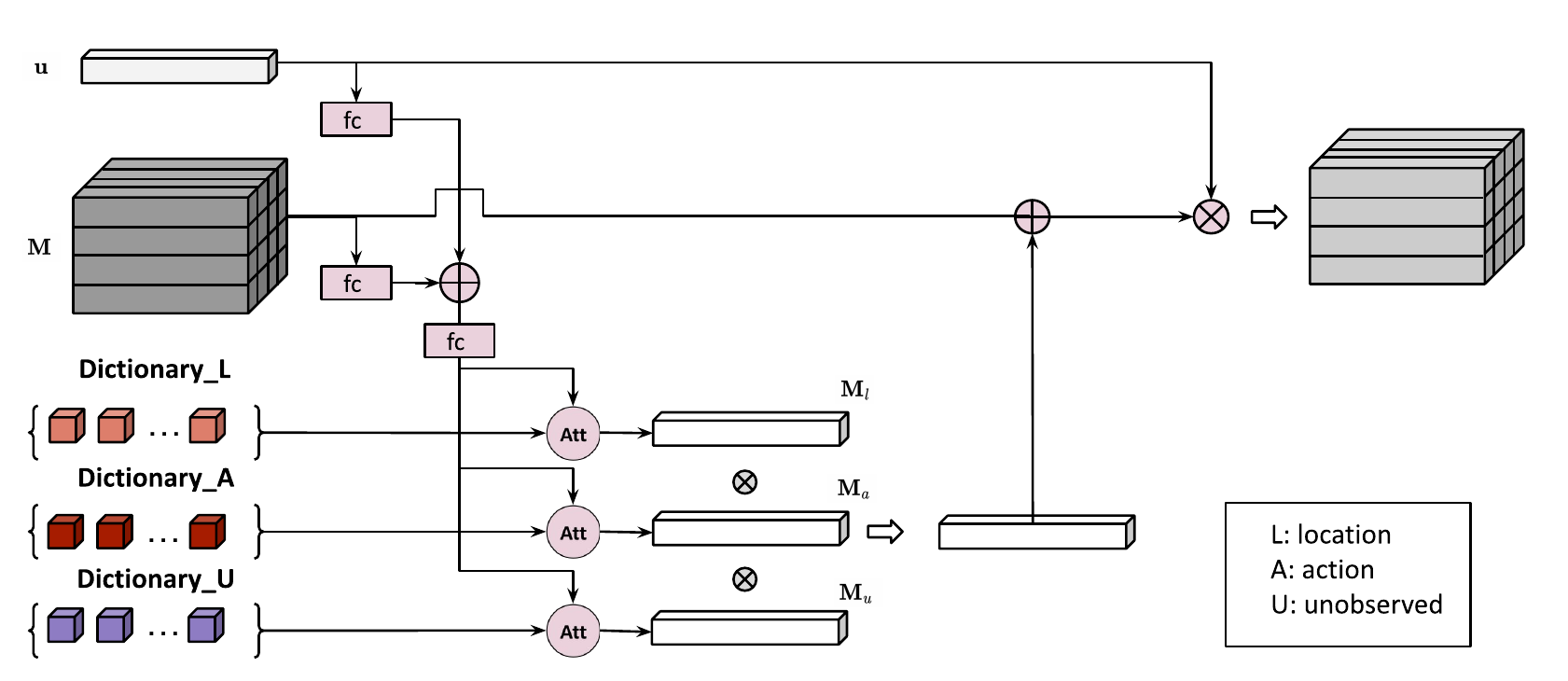}%
\label{fig:deconfounder}
}
% }
\caption{(a) is the causal graph for one single confounder. (b) illustrates the pipeline of multi-branch deconfounder, the dictionary of one single confounder is aggregated by the weights based on the fusion of multimodal inputs.}
\label{fig:deconfounder-graph}
% \end{center}
\end{figure*}

\subsection{Multi-branch Deconfounder} \label{multi-branch-deconfounder}
\textbf{Analysis on Multiple Confounders.}
Inspired by the work~\cite{yang2021deconfounded}, we leverage the structured causal model to analyze the underlying relations among all variables of this TSGV problem. The causal graph which is a directed acyclic graph (DAG) is shown in Fig.~\ref{fig:deconfounder-graph} (a), where the nodes denote the variables and the directed edges denote the relations between nodes. $Q$ is the variable of query, $V$ denotes the video moment and $Y$ is the variable of predicted matching score. For those traditional TSGV models, they train a model to obtain the probabilities $P(Y|Q,V)$ that is conditioned on $Q$ and $V$. However, there may exist a confounder $C$ that has connections with both the multimodal inputs (\ie, $V$ and $Q$) and output scores $Y$. The confounder is harmful since it causes spurious correlation between the inputs and outputs. 

We further investigate the characteristics of TSGV task and find there may exist multiple confounders. Some of the confounders are observable, \eg, the location variable $L$~\cite{yang2021deconfounded}. Since the location information is naturally encoded in the moment representations while we can also use the moment location distribution priors shown in Fig.~\ref{fig:resplit_dataset} to perform moment predictions. Moreover, the action variable $A$ could also be the confounder. The activity concepts implicitly exist in the inputs of video moments and queries while the model could also predict the matching score with only the action label. For example, it can localize on a short moment at the beginning of the video when seeing action ``open'' based on the action-conditioned moment annotation distribution shown in Fig.~\ref{fig:resplit_dataset_action}. Besides, some of the confounders (denoted as $U$) are not observable, such unobserved confounders should also be taken into consideration. Therefore, the do-calculus operation for intervening multiple confounders should be:
\begin{equation}
    P(Y|do(Q,V)) = \sum_{l}P(l)\sum_{a}P(a)\sum_{u}P(u)\cdot P(Y|Q,V,l,a,u) \,.
\end{equation}
Here, we assume that all the confounder variables are independent of each other.

\textbf{Implementation of Base Model.}
After obtaining the 2D temporal moment feature $\mathbf{M}$ and gated fine-grained query feature $\mathbf{u}$, the probabilities $P(Y|Q,V)$ without do-calculus can be learned by:
\begin{equation}
    P(Y|Q,V) \approx \sigma(\textbf{W}^T(\Phi_{conv}(\mathbf{u} \odot \mathbf{M})))
    \,.
\end{equation}
Here, the moment features are fused with the broadcasting query feature via Hadamard product. Then such multimodal representations are fed into the temporal convolutional network $\Phi_{conv}$, followed by a fully connected layer with learnable matrix $\textbf{W}^T$ and the Sigmoid function $\sigma(\cdot)$ to get the final 2D temporal matching scores. 

\textbf{Implementation of Multi-branch Deconfounder.}
As shown in Fig.~\ref{fig:deconfounder-graph} (b), we consider getting three confounders $L$, $A$ and $U$ intervened as the multi-branch deconfounder. Each confounder is represented by a dictionary of enumerable elements. Specifically, we implement such intervention by adding a weighted embedding of all elements in the dictionary for each query-moment pair. More concretely, we assign the dictionary of location $L$ with the 2D position encodings which is the same as the position reconstruction module (Section ~\ref{sec:position-reconstruct}), and we initiate the dictionary of action $A$ with the corresponding word embeddings of limited top-frequency action labels. The unobserved confounder $U$ can be represented by learnable dictionary embeddings of a fixed size. In order to get all confounders intervened at the same time, the weighted representations of multiple confounders are subsequently fused by element-wise multiplication to achieve multi-branch de-confounding. $P(Y|do(Q,V))$ can be approximated as:
\begin{equation}
    P(Y|do(Q,V)) \approx \sigma(\textbf{W}^T(\Phi_{conv}(\mathbf{u} \odot (\mathbf{M} +  \\
    \mathbf{M}_l \odot \mathbf{M}_a \odot \mathbf{M}_u))))
    \,,
\end{equation}
where the effects of multiple confounders are implemented by integrating all the weighted 2D embedding $\mathbf{M}_k \in \mathbb{R}^{N^v \times N^v \times d^v}, k \in \{l,a,u\}$, and then adding such integrated embedding to $\textbf{M}$ (\cf, Fig.~\ref{fig:deconfounder-graph} (b)). Each $\mathbf{M}_k$ with $ k \in \{l, a, u\}$ denotes the effect of any confounder belonging to $\{L, A, U\}$, which is the weighted average of all elements within the dictionary $\mathbb{E}_k[h_{qv}(k)]$.  $\mathbb{E}_k[h_{qv}(k)]$ can be computed with the multi-head attention module~\cite{vaswani2017attention} whose query is the fusion of $\textbf{M}$ and $\textbf{q}^u$. In other words, the attention weight of each element within the dictionary is determined by each query-moment pair. Specifically, $\mathbf{M}_k$ can be defined as:
\begin{equation}
    \begin{split}
    \mathbf{M}_k &= \mathbb{E}_k[h_{qv}(k)] = \textrm{Concat}(\widetilde{\mathbf{A}}_1,\dots,\widetilde{\mathbf{A}}_H) \,, \\ 
    \widetilde{\mathbf{A}}_i &= \left[\textrm{Softmax}(\frac{\mathbf{Q}_i\mathbf{D}_{ki}^T}{\sqrt{d^H}}) \mathbf{D}_{vi} \right] \quad i=1,\dots,H\,, % (L^q, d^H) x (d^H, N^k) x (N^k, d^H) 
    \end{split}
\end{equation}
where $H$ is the head number and $d^H=\dfrac{d^v}{H}$ is the dimension of each subspace. $\textbf{D} \in \mathbb{R}^{N^k \times d^v}$ represents the dictionary containing $N^k$ elements. 
And $\textbf{D}_k = \textbf{D} \textbf{W}_1$, $\textbf{D}_v = \textbf{D} \textbf{W}_2$ with learnable parameters $\textbf{W}_1$, $\textbf{W}_2 \in \mathbb{R}^{d^v \times d^v}$.
% Both $\textbf{D}_1$ and $\textbf{D}_2 \in \mathbb{R}^{N^k \times d^v}$ are obtained by $D$.
% \textcolor{red}{Both $\textbf{K}$ and $\textbf{V} \in \mathbb{R}^{N^k \times d^v}$ are the linear transformation of $N^k$ embeddings within the dictionary into the dimension of $d^v$. (Confused! Rewrite!)}  
The query for multi-head attention is $\textbf{Q} = \textrm{FC}_q(\textrm{FC}_u(\textbf{q}^u) + \textrm{FC}_m(\textbf{M}))$, where $\textrm{FC}_q$, $\textrm{FC}_u$, $\textrm{FC}_m$ are all the fully connected layers with learnable parameters $\in \mathbb{R}^{d^v \times d^v}$. Note that $\textbf{Q}$ is flatten to $\mathbb{R}^{L^q \times d^v}$ for subsequent computation, where $L^q = N^v \times N^v$. Then $\textbf{D}_k$ is equally divided into $H$ parts $\{\textbf{K}_i\}_1^H \in \mathbb{R}^{N^k \times d^H}$ along the feature dimension, so do $\textbf{D}_v$ and $\textbf{Q}$. 

\subsection{Learning Objectives}
Besides the reconstruction loss $L_{recon}$, we use the scaled ground-truth IoU in \cite{zhang2020learning} as the binary cross entropy loss:
\begin{equation}
    \mathcal{L}_{bce} = \sum_{i=1}^N y_i\log s_i + (1-y_i)\log (1-s_i)\,,
\end{equation}
where $y_i$ is the scaled IoU score and $s_i$ is the predicted matching score. The final learning objectives are defined as:
\begin{equation}
    \mathcal{L} = \mathcal{L}_{bce} + \lambda  \mathcal{L}_{recon}\,,
\end{equation}
where $\lambda$ is the hyperparameter.
\section{Experiments}
In this section, we conduct a series of experiments to validate the effectiveness of new evaluation protocols and our proposed debiasing framework. 

\subsection{Implementation Details}
For benchmarking existing methods, we used their open-sourced codes and claimed hyperparameters to train the models with our proposed data splits. The models were validated by the iid set and tested by both the test-iid and test-ood sets. For fair comparisons, we uniformly used pre-trained I3D~\cite{carreira2017quo} features for Charades-CD and C3D~\cite{tran2015learning} features for ActivityNet-CD as video encoding. For query encoding, we used GloVe~\cite{pennington2014glove} to embed the words.

For our debiasing framework, we followed \cite{zhang2020learning} to use three-layer uni-directional LSTM to sequentially encode the queries and adopted max-pooling strategy for moment feature extraction. For both datasets, all of the hidden sizes (\ie, $d^v$, $d^h$ and $d^p$) were set to $512$ and the number of sampled clips was set to $16$. The head number $H$ was set to $4$. The number of stacked convolutional layers for predicting matching scores was set to $4$ with kernel size of $5$. During training process, the batch size and non maximum suppression threshold were set to $64$ and $32$ for Charades-CD and ActivityNet-CD, respectively, and hyperparameter $\lambda$ was $1$. We used Adam optimizer~\cite{kingma2014adam} with learning rate of $1\mathrm{e}{-4}$.
% the batch size and non maximum suppression threshold are set to 64/0.45 and 32/0.5 for Charades-CD and ActivityNet-CD, respectively. (tmp)

% For all these SOTA methods, we use the public available official implementations to get the temporal grounding results. The results of the proposed test-iid and test-ood sets on two datasets come from the same model finetuned on the val set. For more fair comparisons, we have unified the feature representations of the videos and sentence queries. To cater for most of TSGV methods, we use I3D feature~\cite{carreira2017quo} for the videos in dataset Charades-STA (Charades-CD), and C3D feature~\cite{tran2015learning} for the videos in dataset ActivityNet Captions (Activity-CD). Each word in the query sentences is encoded by a pretrained GloVe~\cite{pennington2014glove} word embedding.
% We leverage the public implementations of all the above existing models to get their temporal grounding results. Meanwhile, for a fair comparison, we also unify the encoding of video and sentence (word) features used by each method. Specifically, to cater to most methods, I3D~\cite{carreira2017quo} features are adopted for Charade-STA and C3D~\cite{tran2015learning} features are adopted for ActivityNet Captions. Each word in the query sentence are encoded by the Glove \cite{pennington2014glove} word representation.

\subsection{Performance Comparisons on the Original and Proposed Data Splits}
% \noindent\textbf{Performance Comparisons on Charades-CD and ActivityNet-CD.}
% We compare our approach with existing methods under new evaluation protocols. 
To evaluate the generalization ability of existing methods and demonstrate the difficulty of the newly proposed splits (\ie, Charades-CD and ActivityNet-CD), we compared the performance of two simple baselines and nine representative SOTA methods. In general, we can group all methods into following categories:
% To evaluate the generalization ability of existing methods and demonstrate the difficulty of the newly proposed splits (\ie, Charades-CD and ActivityNet-CD), we compare the performance of two simple baselines and nine representative state-of-the-art methods. Specifically, we can categorize these methods into the following groups: 
\begin{itemize}
\vspace{-0.1em}

\item \textit{Non-deep methods}: Non-deep methods contain two simple baselines without training. The first one is the \textbf{Bias-based} method, which uses the Gaussian kernel density estimation to fit the moment annotation distribution, and randomly samples several locations based on the fitted distribution as the final moment predictions. The second one is the \textbf{PredictAll} method, which directly predicts the whole video as the final moment predictions.

\item \textit{Two-Stage methods}: Cross-modal Temporal Regression Localizer (\textbf{CTRL})~\cite{gao2017tall}, and Attentive Cross-modal Retrieval Network (\textbf{ACRN})~\cite{liu2018attentive}.

\item \textit{End-to-End methods}: Attention-Based Location Regression \\ (\textbf{ABLR})~\cite{yuan2019find}, 2D Temporal Adjacent Network (\textbf{2D-TAN})~\cite{zhang2020learning}, Semantic Conditioned Dynamic Modulation (\textbf{SCDM})~\cite{yuan2019semantic}, and Dense Regression Network (\textbf{DRN})~\cite{zeng2020dense}.

\item \textit{RL-based method}: Tree-Structured Policy based Progressive Reinforcement Learning (\textbf{TSP-PRL})~\cite{wu2020tree}.

\item \textit{Weakly-supervised method}: Weakly-Supervised Sentence Localizer (\textbf{WSSL})~\cite{duan2018weakly}.

% \item \textit{\textcolor{red}{debiasing temporal grounding methods (Maybe change the name, and delete 2D-TAN here.)}}: Both of \textbf{TCN-DCM}~\cite{yang2021deconfounded} and our method use the end-to-end framework 2D-TAN as the base model for debiasing.

\end{itemize}

\begin{figure*}[!t]
	\centering
	\includegraphics[width=1.0\textwidth]{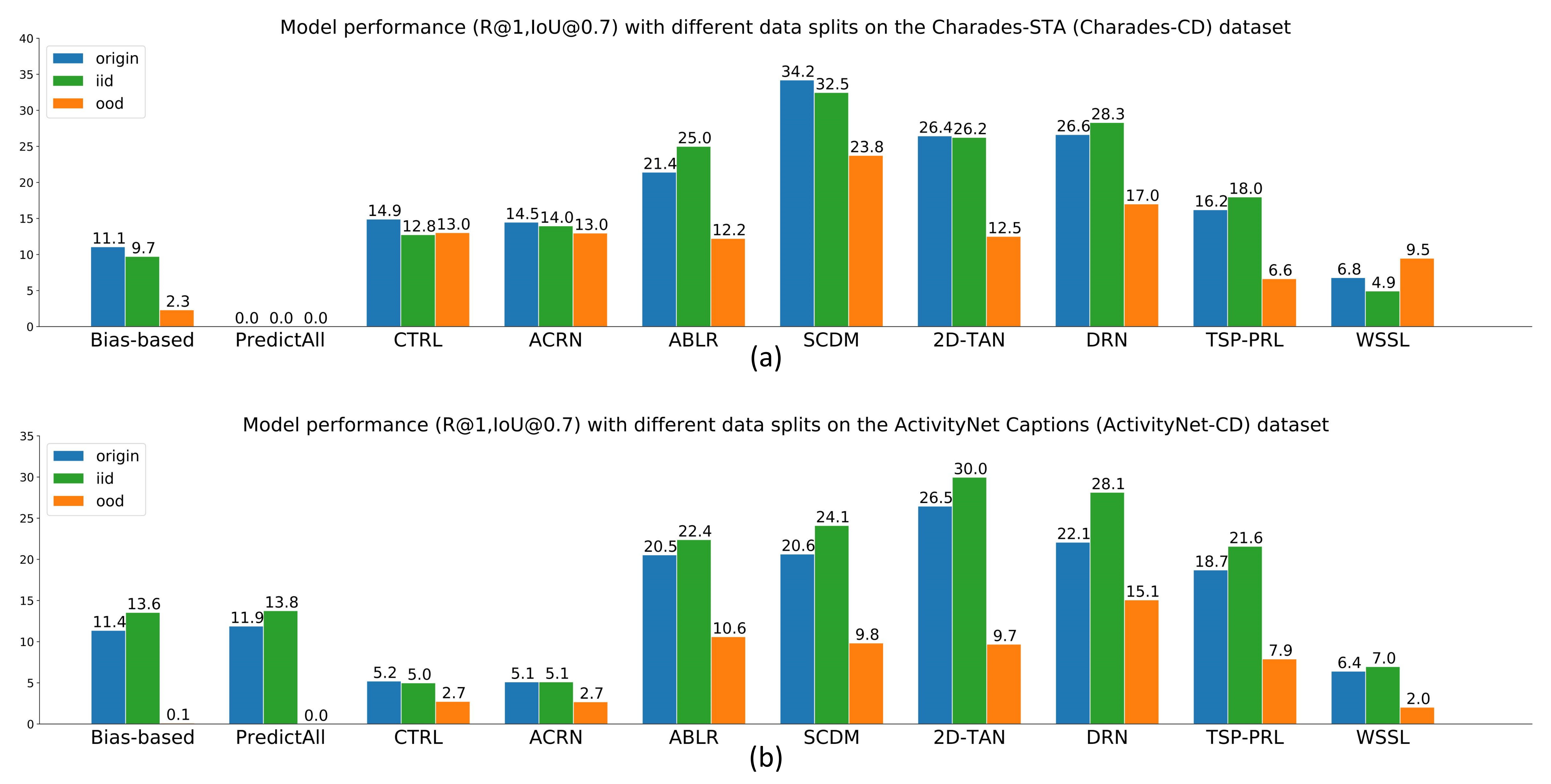}
	\caption{Performances (\%) of SOTA TSGV methods on the test set of original splits (Charades-STA and ActivityNet Captions) and test sets (test-iid and test-ood) of proposed splits (Charades-CD and ActivityNet-CD). We use metric R@$1$,IoU@$0.7$ in all cases.}
	\label{fig:ood_origin_compare}
\end{figure*}

We report the performance of all mentioned TSGV methods with metric     ``dR@$1$,IoU@$0.7$'' in Fig.~\ref{fig:ood_origin_compare}. We can observe that almost all methods have a significant performance gap between the test-iid and test-ood sets, \ie, these methods are prone to over-relying on the moment annotation biases, and fail to generalize to the OOD test. Meanwhile, the evaluation results on the original test set and the proposed test-iid set are relatively close, which shows that the moment distribution of the test-iid set is similar to the majority of the whole dataset. More detailed experimental result analyses are provided in the following:
% We report the performance of all mentioned TSGV methods with metric R@$1$,IoU@$0.7$ in Table~\ref{tab:perf-new-metric}. We can observe that almost all methods have a significant performance gap between the test-iid and test-ood sets, \textit{i.e.}, these methods always over-rely on the moment annotation biases, and fail to generalize to the OOD test. Meanwhile, the performance results on the original test set and the proposed test-iid set are relatively close, which shows that the moment distribution of the test-iid set is similar to the majority of the whole dataset. We provide more detailed experimental result analyses in the following:
% Figure \ref{fig:ood_origin_compare} shows the R@1,IoU@0.7 performance of different methods on the original and our reorganized data splits. We can see that on both of the datasets, the performance of most of the tested methods drop significantly on the test-ood set compared to the original test and test-iid sets. The evaluated results on the original and test-iid sets are relatively close to each other, because their temporal annotation distributions are similar to those in the training set.

\textbf{Non-deep Methods.} The Bias-based method that only exploits the annotation biases of the training set is apparently unable to perform well after changing the moment annotation distributions in test splits. Statistically, its performance on ActivityNet-CD heavily degrades from 13.6\% of the test-iid set to 0.1\% of the test-ood set. As for the PredictAll method, since all the ground-truth moments in Charades-CD 
account for less than 50\% range of the whole videos, simply taking the entire video as the prediction will inevitably lead to ``R@$1$,IoU@$0.7$'' of 0.0 on all test splits. The test samples in ActivityNet-CD are much longer, so the PredictAll method can achieve high results of 11.9\% and 13.8\% on the original test set and new test-iid set, respectively. However, the longer moments are excluded in the test-ood set, thus the performance decreases to 0.0 as well.

\textbf{Two-Stage Methods.} We find that the two-stage methods (\ie, CTRL and ACRN) are less sensitive to the domain gaps between the test-iid and test-ood sets. This is because they utilize a sliding-window strategy to obtain moment candidates, and match these moment candidates with each query individually. In this way, all moment candidates without specific positional attributes are treated equally, thus the moment annotation distribution has less effect on the evaluation results. It is observed that the performance of the test-iid and test-ood sets on Charades-CD are competing while the OOD performance presents a more obvious drop on ActivityNet-CD. The primary reason for this observation is that the moment candidates have more chance to hit the Charades-CD ground-truth moments, which take up a longer percentage of the entire videos (\cf, Fig.~\ref{fig:resplit_dataset} (a)). Despite the less sensitivity against the annotation biases, the performances of these two-stage methods are still far behind those of the SOTA methods from other categories.
% We find that the two-stage methods (\ie, CTRL and ACRN) are less sensitive to the domain gaps between the test-iid and test-ood sets. This is due to that they use a sliding-window strategy to retrieve video moment candidates, and compare these moment candidates with each query sentence individually. In this manner, all moment candidates are independent to the overall video contents, and the moment annotation distributions have less influence on the model performance. We can also observe that the performance of these two methods on the test-ood set of ActivityNet-CD presents a more obvious drop compared to the performance on test-iid set. In contrast, the performance on the test-iid and test-ood sets of Charades-CD are competing. The main reason is that the ground-truth moments in the test-ood set of Charades-CD always occupy a longer range over the whole videos (cf. Fig.~\ref{fig:resplit_dataset} (a), which makes the sliding windows have more chance to hit the ground-truth moments. In summary, although CTRL and ACRN are less sensitive to the moment annotation biases, their grounding performances are still far behind other types of SOTA methods, \eg, SCDM and DRN.

\textbf{End-to-End Methods.} 
For all tested end-to-end methods, we can observe common and significant performance drops on the test-ood set compared to the test-iid set with both two datasets. All of these methods have considerate thoughts about the temporal relations and contextual information of the whole video, since some queries may contain words indicating temporal locations and orders like ``begin'', ``end'', ``first'', ``before'' and ``after'', and some of them intend to model the temporal relations between moments. Unfortunately, although our test-ood split does not break any temporal relations, their OOD performances still drop significantly, which demonstrates that current TSGV methods fail to utilize the visual temporal relation or cross-modal interaction.

\textbf{RL-based Method.} The RL-based method (\ie, TSP-PRL) suffers from obvious OOD performance drops on the test-ood set as well. TSP-PRL adopts IoU between current predicted moment and the ground-truth at each step as the training reward, so the temporal annotations can directly affect the learning process. Therefore, the changing of temporal annotation distributions will inevitably cause the model performance degradation.
% The RL-based method TSP-PRL also suffers from obvious performance drops on the test-ood set compared to the test-iid set. Actually, TSP-PRL adopts IoU between the predicted and ground-truth moment as the training reward in the RL framework. In this case, the temporal annotations directly affect the model learning, and the changes of moment annotation distributions will inevitably influence the model performance.
% The RL-based method TSP-PRL also has obvious performance drop on the test-ood sets of both datasets. Actually, TSP-PRL adopts IoU between the predicted and ground-truth segments as the reward score in its reinforcement learning framework. In this case, the temporal annotations directly affects the model learning, and changing its distribution will inevitably cause the model performance degradation.

\textbf{Weakly-supervised Method.} The evaluation results of the weakly-supervised method WSSL are thought-provoking: it achieves higher performance on test-ood set compared to test-iid set in Charades-CD, but results of these splits in ActivityNet-CD are exactly the reverse. One key finding after investigating the grounding results is that the normalized (start, end) moment predictions of both two re-organized datasets converge on a few certain intervals (\ie, (0, 1), (0, 0.5), (0.5, 1)), which indicates that WSSL does not learn the semantic alignment between the videos and sentences at all. It only speculatively guesses several likely locations instead.
% The results of the weakly-supervised method WSSL are thought-provoking: it achieves better performance on test-ood set compared to test-iid set in Charades-CD, but results of these splits in ActivityNet-CD are exactly the reverse. After carefully checking the predicted moment results, we find that the normalized (start, end) moment predictions on both two datasets converge on several certain predictions (\ie, (0, 1), (0, 0.5), (0.5, 1)). These results indicate that the WSSL method does not learn to align the video and sentence semantics at all. Instead, it only speculatively guesses several possible locations.

% \noindent\textbf{debiasing Temporal Grounding Methods.}\textcolor{red}{Base model with debiasing variants. (Change the bullet name? Maybe di-biased temporal grounding methods)}

\subsection{Our Proposed MDD Framework vs. SOTA Methods}
\begin{table}[!tb]
\centering
\caption{The performance comparison with dR@$1$,IoU@\{$0.5$,$0.7$\} (The \textbf{BOLD} number indicates the best performance and the \underline{UNDERLINE} number indicates the second best one.)}
\label{tab:perf-new-metric}
\begin{adjustbox}{width={0.75\textwidth},totalheight={\textheight},keepaspectratio}%
\begin{tabular}{@{}lcccccccc@{}}
\toprule
\multirow{3}{*}{} & \multicolumn{4}{c}{Charades-CD} & \multicolumn{4}{c}{ActivityNet-CD} \\ \cmidrule(l){2-9} 
 & \multicolumn{2}{c}{test-iid} & \multicolumn{2}{c}{test-ood} & \multicolumn{2}{c}{test-iid} & \multicolumn{2}{c}{test-ood} \\ \cmidrule(l){2-9} 
 & 0.5 & 0.7 & 0.5 & 0.7 & 0.5 & 0.7 & 0.5 & 0.7 \\ \midrule
Bias-based & 16.87 & 9.34 & 5.04 & 2.21 & 19.81 & 12.27 & 0.26 & 0.11 \\ 
PredictAll & 0.00 & 0.00 & 0.06 & 0.00 & 20.05 & 12.45 & 0.00 & 0.00 \\ \midrule
CTRL~\cite{gao2017tall} & 29.80 & 11.86 & 30.73 & 11.97 & 11.27 & 4.29 & 7.89 & 2.53 \\ 
ACRN~\cite{liu2018attentive} & 31.77 & 12.93 & 30.03 & 11.89 & 11.57 & 4.41 & 7.58 & 2.48 \\ 
ABLR~\cite{yuan2019find} & 41.13 & 23.50 & 31.57 & 11.38 & 35.45 & 20.57 & \underline{20.88} & 10.03 \\ 
2D-TAN~\cite{zhang2020learning} & 46.48 & 28.76 & 28.18 & 13.73 & 40.87 & 28.95 & 18.86 & 9.77 \\ 
SCDM~\cite{yuan2019semantic} & 47.36 & 30.79 & \textbf{41.60} & \underline{22.22} & 35.15 & 22.04 & 19.14 & 9.31 \\ 
DRN~\cite{zeng2020dense} & 41.91 & 26.74 & 30.43 & 15.91 & 39.27 & 25.71 & \textbf{25.15} & \textbf{14.33} \\ \midrule
TSP-PRL~\cite{wu2020tree} & 35.43 & 17.01 & 19.37 & 6.20 & 33.93 & 19.50 & 16.63 & 7.43 \\ \midrule
WSSL~\cite{duan2018weakly} & 14.06 & 4.27 & 23.67 & 8.27 & 17.20 & 6.16 & 7.17 & 1.82 \\ \midrule
TCN-DCM~\cite{yang2021deconfounded} & \underline{52.50} & \textbf{35.28} & \underline{40.51} & 21.02 & \underline{42.15} & \underline{29.69} & 20.86 & 11.07 \\ 
MDD~(Ours) & \textbf{52.78} & \underline{34.71} & 40.39 & \textbf{22.70} & \textbf{43.63} & \textbf{31.44} & 20.80 & \underline{11.66} \\ \bottomrule
\end{tabular}
\end{adjustbox}
\end{table}

We also compare our approach to the above methods with the new metrics dR@$1$,IoU@\{$0.5$,$0.7$\}. As shown in Table~\ref{tab:perf-new-metric}, our approach outperforms the base model 2D-TAN with a great gain and has comparable results with another debiasing method TCN-DCM~\cite{yang2021deconfounded}.

For Charades-CD dataset, MDD achieves the best results on both iid-$0.5$ and ood-$0.7$ (iid/ood-$m$ denotes ``dR@$1$,IoU@$m$'' for test-iid/ood set). The performance of MDD on iid-$0.7$ and ood-$0.5$ is slightly lower than the best with 0.57\% and 1.21\%, respectively. These observations indicate that the enhancement of textual and visual features and the causal intervention strategy via multi-branch deconfounder can effectively improve the performance and increase the robustness of moment prediction. For ActivityNet-CD, the absolute gain of MDD against the base model (\eg, 2.76\%/2.49\% on iid-$0.5$/iid-$0.7$) is not as significant as Charades-CD (\eg, 6.30\%/5.95\% on iid-$0.5$/iid-$0.7$) since ActivityNet-CD is more challenging with diverse actions and complex scenarios. But MDD obviously surpasses all methods with test-iid set and get competitive results with test-ood set, which demonstrates that the fine-grained extraction module can better capture the relations of different objects within the queries, and the reconstruction module can obtain more discriminative moment features for further cross-modal matching.

\subsection{Ablation Studies for MDD Framework}
\subsubsection{Model Component Analysis.} 
\npdecimalsign{.}
\nprounddigits{2}
\begin{table}[!tb]
\centering
\caption{The ablation studies of our model on ActivityNet-CD (EM: Enhanced Modalities,  MC: Multi-branch Confounder ($avg_L * avg_U * avg_U$).}
\label{tab:abla-anet}
\begin{adjustbox}{width={0.75\textwidth},totalheight={\textheight},keepaspectratio}%
\begin{tabular}{@{}lcccccccc@{}}
\toprule
\multirow{3}{*}{} & \multicolumn{4}{c}{w/old metric} & \multicolumn{4}{c}{w/new metric} \\ \cmidrule(l){2-9} 
 & \multicolumn{2}{c}{test-iid} & \multicolumn{2}{c}{test-ood} & \multicolumn{2}{c}{test-iid} & \multicolumn{2}{c}{test-ood} \\ \cmidrule(l){2-9} 
 & 0.5 & 0.7 & 0.5 & 0.7 & 0.5 & 0.7 & 0.5 & 0.7 \\ \midrule
base & 46.35 & 31.25 & 21.36 & 10.37 & 40.87 & 28.95 & 18.86 & 9.77 \\ \midrule
base + EM & 47.89 & 32.94 & 22.75 & 11.73 & 42.48 & 30.69 & 20.35 & 11.08 \\ \midrule
base + EM + MC & \textbf{49.03} & \textbf{33.72} & \textbf{23.19} & \textbf{12.33} & \textbf{43.63} & \textbf{31.44} & \textbf{20.80} & \textbf{11.66} \\ \bottomrule
\end{tabular}
\end{adjustbox}
\end{table}
\npnoround 
We investigate the effects of each component in our proposed MDD model, including the modality enhancement module and causality-based multi-branch deconfounder module. As shown in Table~\ref{tab:abla-anet}, the \textbf{base} model is implemented by 2D-TAN, and a visible gain can be observed in the \textbf{base + EM} model after improving the representations of two modalities as described in Section~\ref{feature-extraction}, since the modality enhancement operation does enhance the representation power. And the \textbf{base + EM + MC} model which further includes the multi-branch confounder (Section~\ref{multi-branch-deconfounder}) yields more improvement based on the \textbf{base + EM} model, proving the effectiveness of intervention of multiple confounders.

\subsubsection{Analysis on multi-branch deconfounder.} 

\begin{table}[!tb]
\centering
\caption{Analysis for different combinations of confounders on Charades-CD}
\label{tab:abla-cha}
\begin{adjustbox}{width={0.85\textwidth},totalheight={\textheight},keepaspectratio}%
\begin{tabular}{@{}lcccccccc@{}}
\toprule
\multirow{3}{*}{} & \multicolumn{4}{c}{w/old metric} & \multicolumn{4}{c}{w/new metric} \\ \cmidrule(l){2-9} 
 & \multicolumn{2}{c}{test-iid} & \multicolumn{2}{c}{test-ood} & \multicolumn{2}{c}{test-iid} & \multicolumn{2}{c}{test-ood} \\ \cmidrule(l){2-9} 
 & 0.5 & 0.7 & 0.5 & 0.7 & 0.5 & 0.7 & 0.5 & 0.7 \\ \midrule
base & 50.67 & 30.38 & 31.58 & 14.68 & 46.48 & 28.76 & 28.18 & 13.73 \\ \midrule
MDD-$avg_L$ & 54.56 & 36.70 & 44.22 & 23.28 & 50.44 & 34.81 & 39.64 & 21.78 \\
MDD-$avg_A$ & 55.89 & 36.45 & 45.70 & 23.34 & 51.50 & 34.57 & \textbf{40.85} & 21.77 \\
MDD-$avg_U$ & 57.11 & 35.72 & 43.95 & 22.60 & 52.68 & 33.98 & 39.37 & 21.16 \\ \midrule
MDD-$avg_L * avg_A$ & 57.11 & 36.57 & 42.76 & 21.56 & 52.75 & 34.79 & 38.26 & 20.16 \\
MDD-$avg_L * avg_U$ & 56.38 & \textbf{37.42} & 44.07 & 22.42 & 52.18 & \textbf{35.57} & 39.51 & 21.01 \\
MDD-$avg_A * avg_U$ & 56.38 & 37.30 & 44.66 & 24.08 & 51.91 & 35.39 & 39.99 & 22.50 \\ \midrule
MDD-$avg_L * avg_A * avg_U$ & 56.50 & 36.09 & 42.44 & 21.03 & 52.02 & 34.19 & 37.96 & 19.70 \\
MDD-$avg_L * avg_U * avg_U$ & \textbf{57.23} & 36.57 & \textbf{45.08} & \textbf{24.32} & \textbf{52.78} & 34.71 & 40.39 & \textbf{22.7} \\ \bottomrule
\end{tabular}
\end{adjustbox}
\end{table}

As shown in Table~\ref{tab:abla-cha}, we further explore the impacts of different combinations of confounders to the model performance on the Charades-CD dataset. For example, MDD-$avg_L*avg_A*avg_U$ denotes the multi-branch deconfounder with combining three confounders including location $L$, action $A$ and unobserved variable $U$. Firstly, we consider using only one variable as the confounder.
It can be observed that the performance of MDD-$avg_{L}$ is close to that of MDD-$avg_{A}$, and both of them can surpass the base model with a large gap. This observation demonstrates that introducing the intervention with any confounder (\ie, location, action, unobserved variables) can benefit the model and reduce the influence of the location bias. Then we attempt to increase the number of confounders and the performance gets higher as the amount increases. After many trials we find that the best case to introduce external intervention for unbiased temporal sentence grounding is using the combination of one location variable and two unobserved variables (\ie, MDD-$avg_{L} * avg_{U} * avg_{U}$) as multiple confounders.

\subsection{Performance Gap Between R@1,IoU@m and dR@1,IoU@m}

\begin{wrapfigure}{r}{0.5\textwidth}
	\centering
	\includegraphics[width=0.48\columnwidth]{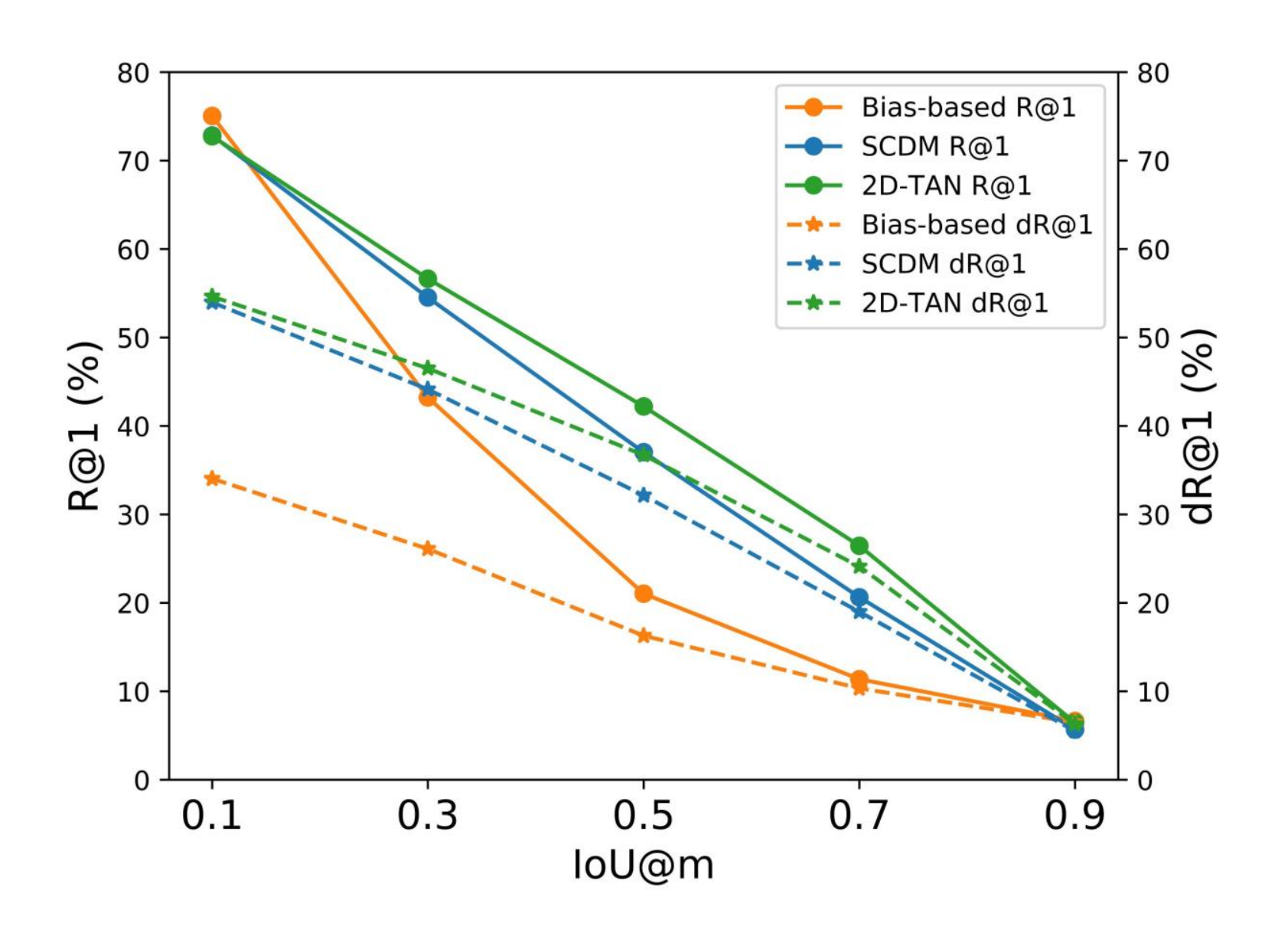}
	\caption{Performance (\%) comparisons of SOTA TSGV methods between original metric (``R@$1$,IoU@$m$'') and proposed metric (``dR@$1$,IoU@$m$''). All results come from the test set of ActivityNet Captions.}
	\label{fig:recall_compare}
\end{wrapfigure}

Fig.~\ref{fig:recall_compare} shows the performance gap between the old and new metric. When the IoU threshold is small, dR@$1$ is much lower than R@$1$, and the gap between them gradually decreases with the increase of IoU threshold. In other words, the performance scores can get discounted by the new metric more heavily under small IoU thresholds, which is able to avoid unreliable evaluation results. For example, it is observed that the simple bias-based method can beat some SOTA methods under the old metric with small IoU threshold of $0.1$, but it cannot outperform others under the new one. This observation further proves the value brought by the proposal of new metrics. 

These results further indicate that recall values under small IoU thresholds are untrustworthy and overrated. Although some moment predictions reach the IoU threshold, they still have a great discrepancy to the ground-truth moments. Instead, our proposed ``dR@$n$,IoU@$m$'' metric can discount the recall value based on the temporal distances between the predicted and ground-truth moment. When the moment prediction meets the larger IoU requirements, the discount effect will be weakened, \ie, the ``dR@$n$,IoU@$m$'' values and ``R@$n$,IoU@$m$'' values will be closer to each other. Therefore, our proposed ``dR@$n$,IoU@$m$'' metric is more stable on different IoU thresholds, and it can suppress some inflating results (such as Bias-based or PredictAll baselines) caused by the moment annotation biases in the datasets. Moreover, the results also reveal that it is more reliable to report the localization accuracy with large IoU thresholds.

\section{Conclusion}
In this paper, we take a closer look at mainstream benchmark datasets for temporal sentence grounding in videos and finds that there exists significant annotation bias, resulting in highly untrustworthy results for evaluating model performance. Therefore, we propose to re-split the datasets so that the location distribution of moment annotation in the training and test sets are different. To alleviate the inflating performance evaluation that is caused by biased datasets as well, we design a new metric to discount the scores considering the temporal distances. The re-organized datasets with the new metric can better monitor current research progress of TSGV. In addition, we design a new debiasing framework to reduce the negative effect caused by the biases from two perspectives: one is to strengthen representations of two modalities, which makes the model easier to learn the semantic alignment between two modalities, and the other is to perform debiasing based on causality, which can both provide good theoretical support and achieve effective debiasing. Experiments show that the newly proposed approach can outperform the base model with a great gap and the evaluation results are also competitive with those of other SOTA models, laying a solid foundation for future research work.

% \section{Acknowledgments}
\begin{acks}
This work is supported by National Natural Science Foundation of China No.62102222 and the National Key Research and Development Program of China No.2020AAA0106300.
\end{acks}

\bibliographystyle{ACM-Reference-Format}
\bibliography{references}

% \appendix

\end{document}